\documentclass{article}
\usepackage{amsmath,amsfonts}
\usepackage{mathtools}
\usepackage{algorithm, float}
\usepackage{algorithmic}
\usepackage{graphicx}
\usepackage{subcaption}
\usepackage{float}

\usepackage{hyperref}
\usepackage{arxiv}

\usepackage{amssymb}
\usepackage{latexsym}
\def\BibTeX{{\rm B\kern-.05em{\sc i\kern-.025em b}\kern-.08em
    T\kern-.1667em\lower.7ex\hbox{E}\kern-.125emX}}
\usepackage{balance}

\begin{document}
\title{Focused Active Learning for Histopathological Image Classification}

\author{ 
{Arne Schmidt} \\
Dep. of Computer Science and AI\\ 
University of Granada\\ 
Granada, Spain \\
\And
{Pablo Morales-Álvarez} \\
Dep. of Statistics and Operations Research\\
University of Granada\\ 
Granada, Spain \\
\And
{Lee A.D. Cooper}\\
Dep. of Pathology\\
Northwestern University\\
Chicago,  USA \\
\And
{Lee A. Newberg} \\
Kitware Inc.\\
Carrboro, USA \\
\And
{Andinet Enquobahrie}\\
Kitware Inc.\\
Carrboro, USA \\
\And
{Aggelos K Katsaggelos}\\
Dep. of Electrical Computer Engineering\\
Northwestern University \\
Evanston, USA\\
\And
{Rafael Molina}\\
Dep. of Computer Science and AI\\ 
University of Granada\\ 
Granada, Spain \\
\thanks{\small{This work was supported by the European Union’s Horizon 2020 research and innovation programme under the Marie Skłodowska Curie grant agreement No 860627 (CLARIFY Project), the US National Institutes of Health National Library of Medicine grant R01LM013523, and the project PID2022-140189OB-C22 funded by MCIN / AEI /
10.13039 / 501100011033. PMA acknowledges grant C-EXP-153-UGR23 funded by Consejería de
Universidad, Investigación e Innovación and by ERDF Andalusia Program.}}
}
\maketitle

\begin{abstract}
Active Learning (AL) has the potential to solve a major problem of digital pathology: the efficient acquisition of labeled data for machine learning algorithms.
However, existing AL methods often struggle in realistic settings with artifacts, ambiguities, and class imbalances, as commonly seen in the medical field. 
The lack of precise uncertainty estimations leads to the acquisition of images with a low informative value.
To address these challenges, we propose Focused Active Learning (FocAL), which combines a Bayesian Neural Network with Out-of-Distribution detection to estimate different uncertainties for the acquisition function.
Specifically, the weighted epistemic uncertainty accounts for the class imbalance, aleatoric uncertainty for ambiguous images, and an OoD score for artifacts.
We perform extensive experiments to validate our method on MNIST and the real-world Panda dataset for the classification of prostate cancer.
The results confirm that other AL methods are 'distracted' by ambiguities and artifacts which harm the performance. FocAL effectively focuses on the most informative images, avoiding ambiguities and artifacts during acquisition.
For both experiments, FocAL outperforms existing AL approaches, reaching a Cohen's kappa of $0.764$ with only $0.69\%$ of the labeled Panda data.
\end{abstract}


\section{Introduction}
Artificial Intelligence (AI) methods have obtained impressive results in digital pathology and in some cases, AI models even outperformed expert pathologists in cancer classification \cite{zhang_pathologist-level_2019, hekler_deep_2019, ehteshami_bejnordi_diagnostic_2017}. The hope is that AI can make the diagnosis more accurate, objective, reproducible, and faster in the future \cite{dimitriou_deep_2019}. 

To achieve this goal, trained, specialized AI models for each subtask are required, for example for the quantification of tumor-infiltrating lymphocytes in lung cancer \cite{shvetsov_pragmatic_2022}, 
metastasis detection of breast cancer in lymph nodes \cite{ehteshami_bejnordi_diagnostic_2017, schmidt_efficient_2022} or Gleason grading of prostate cancer \cite{bulten_artificial_2022, otalora_semi-weakly_2020}.  
Openly available, labeled datasets are limited to certain subtasks and for many future applications, the aggregation of large amounts of labeled data remains challenging because the annotation requires medical experts. This makes the labeling process time-consuming and expensive. 
A common approach to labeling is to divide a region or Whole Slide Image (WSI) into small patches that are individually labeled \cite{dimitriou_deep_2019}. The model is then trained to make local patch-level predictions that can be aggregated for the final diagnosis. The problem with supervised deep learning methods is the need for large amounts of detailed (patch-level) annotations for training to obtain a satisfying predictive performance.
To alleviate this burden, \emph{semi-supervised learning} \cite{li_em-based_2018, marini_semi-supervised_2021, lu_semi-supervised_2020, schmidt_efficient_2022, otalora_semi-weakly_2020} and \emph{multiple instance learning} \cite{campanella_clinical-grade_2019, chikontwe_multiple_2020, li_dual-stream_2021} have become major fields of interest in the recent years. Despite the importance of these fields, there is another approach to efficiently handle labeling resources with several advantages over semi-supervised and multiple instance learning.

\emph{Active Learning (AL)} describes machine learning methods that actively query the most informative labels. In the AL setting, the AI model starts training with a small set of labeled images and iteratively selects images from a large pool of unlabeled data. These selected images are  labeled in each iteration by an 'oracle,' in our application a medical expert.
AL has several advantages over semi-supervised and multiple instance learning: 
(i) The model training and dataset creation go hand-in-hand. 
The performance of the model is constantly monitored to assess if the collected labeled data is enough - or if more labeled data is needed to reach the desired performance. 
(ii) The model looks for the most informative images  automatically in the acquisition step. In semi-supervised learning in comparison (and other methods that require labeling), finding these informative, salient images requires a lot of manual searching. 
(iii) AL is very data-efficient while multiple instance learning often requires large datasets to compensate for missing instance labels \cite{campanella_clinical-grade_2019}.
Furthermore, the AL model can be trained to make accurate local (patch-level) predictions while multiple instance learning models often do not provide those and therefore lack explainability. In other cases, the multiple instance learning setting is not applicable at all because the global classes differ from the local classes. As an example consider images with global labels 'cat' and 'dog'. It would not be possible to train a multiple instance learning model for classifying 'paws' and 'ears' at the image-patch level because this information simply can not be deducted from the global labels. AL models can be trained to make any kind of local predictions with reduced labeling effort.

\noindent 
\textbf{Related work}
In AI research, different AL strategies have been proposed to determine the most informative images. Early approaches used the uncertainty estimation of support vector machines \cite{joshi_multi-class_2009}, Gaussian processes \cite{li_adaptive_2013} or Gaussian random fields \cite{zhu_combining_2003} to rate the image informativeness. 
With the rise of deep learning, the focus shifted to Bayesian Neural Networks (BNNs) for AL \cite{gal_deep_2017}, which was adapted several times for histopathological images \cite{raczkowski_ara_2019,carse_active_2019,meirelles_effective_2022}. This approach has the advantage of a probabilistic uncertainty estimation which is not only used for acquisition, but it is also crucial for diagnostic predictions in medical applications. BNNs allow the application of several different uncertainty-based acquisition functions, such as \emph{BALD} \cite{houlsby_bayesian_2011}, \emph{Max Entropy} \cite{raczkowski_ara_2019, shannon_mathematical_1948}, and \emph{Mean Std} \cite{alex_kendall_bayesian_2017}. 
Other publications focus on the user interface and server application of AL \cite{lee_interactive_2021, maree_collaborative_2016} rather than the AL model itself.
In the existing literature, the uncertainty estimation is often only used to determine the amount of new information in each image. We extend this idea by using complementary uncertainty measures to \emph{avoid} labeling uninformative, ambiguous, or artifactual images.
In digital pathology, several data-related challenges like artifacts, ambiguities, and the typical huge class imbalance hinder the application of AL (see "Problem analysis" paragraph below). Our proposed method tackles these problems successfully by precise uncertainty estimations which leads to improved performance. 

BNNs are not only of interest for the AL acquisition, their capacity to estimate the predictive uncertainty is highly important in safety-critical areas like medicine \cite{kwon_uncertainty_2020} or autonomous driving \cite{alex_kendall_bayesian_2017}. The uncertainty estimation helps to distinguish confident predictions from risky ones. In our case, we aim to decompose uncertainty into \emph{epistemic uncertainty} and \emph{aleatoric uncertainty} describing the model and data uncertainty, respectively \cite{der_kiureghian_aleatory_2009}. Epistemic uncertainty describes uncertainty in model parameters that can be reduced by training with additional labeled data.
Therefore it can serve as a measure of informativeness in the active learning process. Unfortunately, the epistemic uncertainty is not only high for informative, in-distribution images, but also for OoD images. In fact, epistemic uncertainty has recently been used explicitly for OoD detection \cite{xiao_wat_2019, mukhoti_deep_2021, nguyen_out_2022}. 
Aleatoric uncertainty describes irreducible uncertainty in the data due to ambiguities that cannot be improved with additional labeling. 
Studies have shown that training with ambiguous data can harm the performance of the algorithm considerably if not taken into account \cite{gao_deep_2017, bernhardt_active_2022}. In the Panda challenge, label noise associated with the subjective grading assigned by pathologists was considered to be a major problem \cite{bulten_artificial_2022}. 

To estimate these uncertainties with BNNs, Kendall \textit{et al.}\cite{kendall_what_2017} proposed a network with two final probabilistic layers, corresponding to the two uncertainty measures. A theoretically sound, more stable, and efficient approach (relying on a single probabilistic layer) was proposed by Kwon \textit{et al.}\cite{kwon_uncertainty_2020}. We base our BNN for uncertainty estimations on the latter method due to the mentioned advantages. In Section \ref{subsec:exp_panda} we outline how the uncertainty estimations can be interpreted in the context of clinical applications like pathology.

To avoid acquiring image patches with artifacts, we apply OoD detection. Commonly, OoD data refers to data that originates from a different distribution than the training data (in-distribution) \cite{sun_out--distribution_2022}. In the context of AL and pathology, we define the in-distribution as the distribution of patches containing (cancerous or non-cancerous) tissue. All the images with artifacts (such as pen markings, tissue folds, blood, or ink) \cite{kanwal_devil_2022} will be considered OoD. These artifacts are inevitable in real-world data and there are several reasons to exclude them from the distribution of interest for acquisition: (i) It is impossible to learn all possible artifacts explicitly due to their wide variability. We argue that a model should reliably classify tissue and predict a high uncertainty for everything it does not know.  (ii) It harms the performance of AL algorithms to acquire  images with artifacts, as we show empirically in Section \ref{sec:experiments}.  (iii) The model should focus on learning what \emph{is} cancerous instead of everything that \emph{is not} cancerous. By learning cancerous patterns it automatically learns what is not cancerous (everything else).

In OoD detection, early methods used the depth \cite{johnson_fast_1998, ruts_computing_1996} or distance \cite{knorr_algorithms_1998, knorr_finding_1999} of datapoints, represented by low-dimensional feature vectors.
With the rise of deep learning, OoD metrics were often applied to the features extracted by a deep neural network \cite{abati_latent_2019, sun_out--distribution_2022, lee_simple_2018}. In this line with previous research, we utilize extracted feature vectors and implement a density-based OoD scoring method \cite{breunig_lof_2000} to detect artifacts in the data.

\noindent 
\textbf{Problem Analysis}
Although AL has a huge potential for digital pathology, we analyze several challenges that hinder its application in practice:
\begin{itemize}
    \item Medical imaging problems like pathology often have a high class imbalance. For example, in prostate cancer grading, the highest Gleason patterns may be underrepresented which needs to be taken into account during acquisition. Other AL algorithms treat each class equally and are not able to acquire a sufficient number of images of this underrepresented class in our experiments (Section \ref{sec:experiments}).
    \item Many patches are ambiguous. There may be patches for which even subspecialists disagree on their label, or patches containing multiple classes. Assigning labels to these patches is difficult and may be detrimental to the quality of the dataset and the algorithm's performance. This not only slows the labeling process down, but it can also add noise to the training data as only one label per patch is assigned. In fact, label noise associated with the subjective grading assigned by pathologists was considered one key problem in the Panda challenge \cite{bulten_artificial_2022}.
    \item WSIs can contain many different artifacts, such as pen markings, tissue folds, ink, or cauterized tissue. Existing AL algorithms often assign a high informativeness to these patches although they do not contain important information for model training, as we show empirically in the experimental section \ref{sec:experiments}.
\end{itemize}
We want to stress that similar problems of class imbalance, ambiguities, and artifacts are present in many other medical imaging applications, such as CT scans for hemorrhage detection \cite{wu_combining_2021}, dermatology images for skin cancer classification \cite{esteva_dermatologist-level_2017} or retinal images for the detection of retinopathy \cite{gulshan_development_2016}.

\noindent 
\textbf{Contribution}
To address these challenges we propose Focused Active Learning (FocAL), a probabilistic deep learning approach that focuses on the underrepresented malignant classes while ignoring artifacts and ambiguous images. More specifically, we combine a Bayesian Neural Network (BNN) with Out of Distribution (OoD) Detection to estimate the three major elements of the proposed acquisition function. The \emph{weighted epistemic uncertainty} rates the image informativeness, taking the class imbalance into account. The \emph{aleatoric uncertainty} is used to avoid ambiguous images for acquisition. The \emph{OoD score} helps to ignore outliers (like artifacts) that do not contribute information for the classification of tissue. We show empirically that these precise uncertainty estimations help to focus on labeling salient, informative images while other methods often fail to address this realistic data setting.

The article is structured as follows. We outline the theory of the proposed model, including the BNN and OoD  components of the acquisition function in Section \ref{sec:method}. In Section \ref{sec:experiments}, we perform an illustrative MNIST experiment to analyze the behavior of existing AL approaches when artifacts and ambiguities are present. Furthermore, we demonstrate that each of our model components works as expected to avoid acquiring images with ambiguities and artifacts, overcoming the problems of the existing approaches. For the Panda prostate cancer datasets we perform an ablation study about the introduced hyperparameters, analyze the uncertainty estimations, and report in the final experiments that our method can reach a Cohen's kappa of $0.763$ with less than $1\%$ of the labeled data ($4400$ labeled image patches). Finally, in Section \ref{sec:conclusions} we conclude our article and give an outlook of future research.

\section{Methods} \label{sec:method}
\begin{figure*}[!t]
     \centering
     \includegraphics[trim={0.0cm 0.0cm 8cm 0.0cm},width=0.8\linewidth]{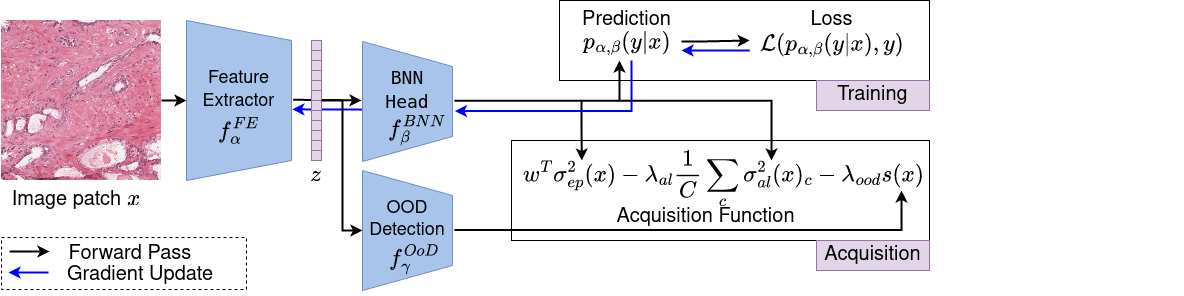}
     \caption{
    Model overview of the proposed FocAL method. It consists of three main components: feature extractor, BNN, and OoD detection (blue boxes). The figure shows how the different components are combined for training and acquisition.} \label{fig:model_overview}
\end{figure*}

Here we describe the three elements of FocAL: the feature extractor, the Bayesian Neural Network, and the Out-of-distribution score. The final paragraph of this section outlines the acquisition function and algorithm of the novel FocAL method. An overview of the model components is depicted in Figure \ref{fig:model_overview}.

\noindent 
\textbf{Active Learning (AL)}
In AL we assume that at the beginning a small set of labeled data $\mathcal{D}_\mathit{train}=\{x_i, y_i\}_{i=1,..,N}$ of images $x_i$ and labels $y_i$ and a pool of unlabeled data $\mathcal{D}_\mathit{pool}$ is available. We assume each $y$ to be a $C$-dimensional, one-hot encoded vector, where $C$ stands for the number of classes. A machine learning model $\mathcal{M}$ trains with a labeled set $\mathcal{D}_\mathit{train}$ and then chooses a subset $\mathcal{A} \subset \mathcal{D}_\mathit{pool}$ of unlabeled images to be labeled (by a specialist, such as a pathologist in the given application). The choice is made with the help of an \emph{acquisition function} $a(x, \mathcal{M})$ which estimates the informativeness of each image $x$.  Probabilistic Active Learning
The images with the highest acquisition scores are labeled and added to $\mathcal{D}_\mathit{train}$. Then, the model is retrained with the updated training set. This acquisition step is repeated iteratively such that the model performance increases while more and more labeled data is aggregated.\\
\noindent 
\textbf{Feature Extraction} 
The feature extractor $f_\alpha^{\mathrm{FE}}$, with model parameters $\alpha$, is the first component of the proposed model. We use a Convolutional Neural Network (CNN) to extract high-level 128-dimensional features $z = f_\alpha^\mathrm{FE}(x)$ from each image patch. The exact architecture of the feature extractor depends on the image data and task, see the implementation details for each experiment in Section \ref{sec:experiments}. The feature extractor is trained during the AL process end-to-end with the BNN by gradient descent, see Figure \ref{fig:model_overview}. Although training the feature extractor is important for obtaining good final results, the emphasis of this article lies on the development of the BNN and OoD detection which perform the high-level reasoning, as described in the next paragraphs.

\noindent 
\textbf{Bayesian Neural Network (BNN)}
The BNN model $f_\beta^\mathrm{BNN}$, with model parameters $\beta$, allows probabilistic reasoning based on the extracted feature vectors $Z=\{z_i\}$. Note that the feature extractor and BNN head together could also be interpreted as a large, convolutional BNN with Bayesian layers near the output. Previous studies have shown that this combination of deterministic convolutions and Bayesian fully connected layers is the most effective way to introduce Bayesian uncertainty in the AL context \cite{zeng_relevance_2018}. Here, we treat the feature extractor and BNN separately, because the features are also used later for OoD detection. The BNN is not only able to make accurate classification predictions, but can also estimate epistemic and aleatoric uncertainty. These estimated uncertainties will be further described below. They play a crucial role in the proposed acquisition function (see the last paragraph of this section).

The BNN in our model consists of two fully connected layers with 128 units and a final softmax output layer. In comparison to deterministic networks with weight parameters $\omega$, BNNs treat the model weights as random variables with a probability distribution $p(\omega)$. As the true posterior distribution $p(\omega|\mathcal{D}_\mathit{train})$ is intractable, it has to be approximated. 
Following the success of similar approaches in recent studies \cite{gal_deep_2017, alex_kendall_bayesian_2017, kwon_uncertainty_2020}, we use variational inference to approximate the posterior distribution by a tractable variational distribution $q_\beta(\omega)$, where $\beta$ describe the variational parameters of the distribution. Specifically, we define $q$ as a product of independent Gaussian distributions over each model weight, parametrized by mean and variance. 
To approximate the real posterior, the minimization of the KL divergence $KL(q_\beta(\omega)|p(\omega|\mathcal{D}_\mathit{train}))$ is achieved by maximizing the evidence lower bound (ELBO) utilizing the reparametrization trick \cite{kingma_variational_2015}. Gradient descent allows the optimization of the BNN and the feature extractor end-to-end. We denote the ELBO loss function as $\mathcal{L}(p_{\alpha, \beta}(y|x),y)$. The predictive distribution $p_{\alpha, \beta}(y|x)$ is obtained by applying the feature extractor $z = f^\mathrm{FE}_\alpha(x)$ and integrating the BNN through Monte Carlo sampling as: 
\begin{align} \label{eq:class_pred}
    p_{\beta}(y|z) &= \int p(y|\omega,z)q_\beta(\omega) \ d \omega \\
    &\approx \frac{1}{T} \sum_{t=1}^T p(y|z, \omega_t), \nonumber
\end{align}
where we use Monte Carlo sampling by drawing $T$ realizations $\{\omega_t\}_{t=1,..,T}$ of the variational weight distribution. The \emph{argmax} over classes of the vector $p_{\beta}(y|z)$ defines the predicted class. For notational convenience, we will drop the parameters $\alpha$ and $\beta$ when not needed.
We show in Figure \ref{fig:model_overview}  an overview of the forward and backward pass with gradient descent.

\noindent 
\textbf{Bayesian Uncertainty Estimations}
In Addition to the class prediction (eq. \ref{eq:class_pred}), the BNN is able to estimate the uncertainty, measured by the predictive covariance matrix $\text{Cov}_{p(y^*|z^*,Z,Y)}(y^*)$. This variance can be further decomposed into epistemic and aleatoric uncertainty.

\emph{Epistemic uncertainty} (model uncertainty) measures the uncertainty introduced by the model parameters $\omega$ and can be reduced with more labeled training data. A high epistemic uncertainty indicates a high informativeness of a given image. Unfortunately, using only the epistemic uncertainty for acquisition can lead to an unwanted outcome. 
If the data is contaminated with outliers such as artifacts, this can result in acquiring only outliers that do not contribute any value towards learning the classes of interest, as shown empirically in the experimental section \ref{sec:experiments}.

\emph{Aleatoric uncertainty} (data uncertainty) captures the uncertainty inherent in the data due to ambiguities. It can not be reduced with more labeled data. Images with a high aleatoric uncertainty should be avoided during acquisition as the chance of mislabeling (due to ambiguous image content) or inherent data noise is higher for these images. 

There are different possibilities for estimating epistemic and aleatoric uncertainty. 
Here, we follow the approach of Kwon \textit{et al.}\cite{kwon_uncertainty_2020}, which does not require additional parameters, is numerically stable, and has a strong theoretical background. The covariance matrix is decomposed into

\begin{align} \label{eq:bnn_uncertainties}
    \text{Cov}_{p(y^*|z^*,Z,X)}(y^*)
    =& \underbrace{\frac{1}{T} \sum_{t=1}^T \{p(y^* | z^*, \omega_t) -  \hat{p}(y^*|z^*) \}^{\bigotimes 2}}_{ \vcentcolon = \text{epistemic uncertainty}} \\
    &+ \underbrace{\frac{1}{T} \sum_{t=1}^T \text{diag} \{p(y^*|z^*, \omega_t)\} - p(y^*|z^*, \omega_t)^{\bigotimes 2}}_{ \vcentcolon = \text{aleatoric uncertainty}} \nonumber
\end{align}
where $\hat{p}(y^*|z^*) = \frac{1}{T} \sum_{t=1}^T p(y^*|z^*, \omega_t)$, $diag\{p(y^*|z^*, \omega_t)\}$ is the diagonal matrix formed by the vector entries of $p(y^*|z^*, \omega_t)$ in the diagonal, and the outer product $v^{\bigotimes 2}=vv^T$. \\
Note that both epistemic and aleatoric uncertainties are given as $C\times C$ covariance matrices with values of the \emph{uncertainty per class} on the diagonal. We define the C- dimensional vectors of class-wise uncertainties as $\sigma_{ep}^2$ for the epistemic and $\sigma_{al}^2$ for the aleatoric uncertainty.

\noindent 
\textbf{Out-of-Distribution (OoD) Detection} 
For the OoD detection, we use an unsupervised, density-based model $f_\gamma^\mathrm{OoD}$ with parameters $\gamma$ \footnote{Note that the parameters $\gamma$ consist of the locations and densities of the currently labeled feature vectors. These are not model parameters in the strict sense but we follow this notation for coherence.}, based on the extracted features $z$. Instead of having a binary decision (in/out of distribution), we want to score each feature vector of the unlabeled images with a Local Outlier Factor (LOF) \cite{breunig_lof_2000}. The LOF is based on the k-nearest neighbors $N_k(z)$ of a vector $z$ and the \emph{local reachability density} $lrd_k(z)$, a density measure based on the distance to the k-nearest-neighbors.\\
The LOF of a vector $z$ is defined as
\begin{equation} \label{eq:lof}
    \mathrm{LOF}_k (z) = \frac{1}{|N_k(z)|}\sum_{z'\in N_k(z)} \frac{\mathit{lrd}_k (z')}{\mathit{lrd}_k (z)}
\end{equation}
with hyperparameter $k$ which should be set to the minimum amount of expected datapoints in a cluster \cite{breunig_lof_2000}.
Intuitively, the LOF is high if a feature vector lies in a region with a lower density than its neighbors (indicating an outlier). If the region of a feature vector has the same density as its neighbors, its LOF is close to $1$. The upper bound depends on the characteristics of the data, i.e. the distances between feature vectors. Empirically we observed that scaling the LOF by $0.1$ leads to an OoD score that is in same the range as the other uncertainty measures (epistemic and aleatoric uncertainty). Therefore, we define the outlier scoring function as
\begin{equation} \label{eq:ood_score}
    s(x) = 0.1 \ LOF_{k}(f^\mathrm{FE}(x)).
\end{equation}
Note that the scaling factor does not introduce an additional hyperparameter. It is inherently tuned by manipulating the weighting factor $\lambda_{ood}$ in eq. \ref{eq:acquisition_function} for which we perform experiments in section \ref{subsec:exp_panda}. The hyperparameter $k$ can be set to a rough estimate of the minimum number of initial images $x \in \mathcal{D}_\mathit{train}$ that are not affected by ambiguities or artifacts. We set it to $k=10$ for MNIST and $k=50$ for the Panda dataset.

\noindent 
\textbf{Focused Active Learning (FocAL)} 
We propose an acquisition function that combines the uncertainty-based measures and the OoD scoring discussed above: 
\begin{equation} \label{eq:acquisition_function}
    a(x, \mathcal{M}) = \underbrace{w^T \sigma_{ep}^2(x)}_{\text{weighted ep. unc.}} 
    - \lambda_{al} \underbrace{\frac{1}{C} \sum_c \sigma_{al}^2 (x)_c}_{\text{aleatoric unc.}}
    - \lambda_{ood} \underbrace{s(x)}_{\text{OoD score}}
\end{equation}
with the (calculated) class weight vector $w=[w_1, w_2, .., w_C]^T$ (see eq. \ref{eq:class_weights}) and hyperparameters $\lambda_{al}\in \mathbb{R}^+$, $\lambda_{ood}\in \mathbb{R}^+$. The images with the highest scores of the acquisition function are selected for labeling in each step. Each component fulfills a specific task in the acquisition process:\\
\emph{1) Weighted Epistemic Uncertainty} With the BNN we can calculate the informativeness of each unlabeled image measured by the epistemic uncertainty. This measure has a sound theoretical background and a proven track record in practice. The advantage of BNNs is that this approach estimates the epistemic uncertainty for each class independently. Existing approaches based on epistemic uncertainty often just take the sum over all classes. For our proposed model, we want to emphasize the informativeness of underrepresented classes. Therefore, we multiply the epistemic uncertainty of each class with a class-weight $w_c$ which is calculated by
\begin{equation} \label{eq:class_weights}
    w_c = \frac{N_\mathit{train}}{N_{c}*C}
\end{equation}
with $N_\mathit{train}$ being the number of labeled images, $N_c$ being the number of labeled images of class $c$ and $C$ the number of classes.
The class weights are recalculated at each acquisition step, depending on the given label distribution of the current set $D_\mathit{train}$. This allows the algorithm to automatically adjust to class imbalances in $\mathcal{D}_\mathit{train}$.
\\
    \emph{2) Aleatoric Uncertainty} The BNN measures the aleatoric uncertainty of each unlabeled image. We down-weight the informativeness of images based on their aleatoric uncertainty estimate to avoid labeling ambiguous patches. Although in the existing literature the aleatoric uncertainty is described as a measure of data uncertainty, we found that it does not capture data uncertainty for images with a different appearance (OoD). Therefore, an additional measure for the OoD images is necessary.\\
    \emph{3) OoD Score} 
    To avoid the acquisition of outliers we apply an OoD algorithm on extracted image features. We down-weight the informativeness of images with a high OoD score. This allows the network to focus on the in-distribution data and acquire informative image patches.
\\
The active learning procedure is summarized in Algorithm \ref{alg}. \\
After the acquisition steps are completed, the trained models are not only able to give accurate classification predictions for each new test image, but also the epistemic and aleatoric uncertainty and OoD score which is very useful for the pathologist in the diagnostic process. In the regions where all three uncertainty measures are low, the prediction is reliable and the pathologist can trust the classification result.

\begin{algorithm}[tb]
\caption{FocAL algorithm} \label{alg}
 \begin{algorithmic}
 \renewcommand{\algorithmicrequire}{\textbf{Input:}}
 \renewcommand{\algorithmicensure}{\textbf{Output:}}
 \REQUIRE Start training set $\mathcal{D}_\mathit{train}^0$, pool of unlabeled data $\mathcal{D}_\mathit{pool}^0$, models $f_\alpha^\mathrm{FE}, f_\beta^\mathrm{BNN}, f_\gamma^\mathrm{OoD}$, number of acquisition steps $S$.
 \ENSURE  Optimal model parameters $\alpha, \beta, \gamma$; training dataset $\mathcal{D}_\mathit{train}^S$ \\
  \FOR {$s= 0$ to $S$}
  \STATE Train $f_\alpha^\mathrm{FE}, f_\beta^\mathrm{BNN}$ with $\mathcal{D}_\mathit{train}^s$.
  \STATE Predict features $Z_\mathit{train} \gets f_\alpha^\mathrm{FE}(X_\mathit{train})$
  \STATE Update $f_\gamma^\mathrm{OoD}$ with $Z_\mathit{train}$
  \STATE Predict features $Z_\mathit{pool} \gets f_\alpha^\mathrm{FE}(X_\mathit{pool})$
  \STATE Estimate unc. $\sigma_{ep}^2(X_\mathit{pool}), \sigma_{al}^2(X_\mathit{pool}) \gets f_\beta^\mathrm{BNN}(Z_\mathit{pool})$
  \STATE Estimate OoD scores $s(X_\mathit{pool}) \gets f_\gamma^\mathrm{OoD}(Z_\mathit{pool})$ (eq. \ref{eq:ood_score})
  \STATE Select acq. set $A^s$ with $a(X_\mathit{pool}, \{f_\alpha^\mathrm{FE}, f_\beta^\mathrm{BNN}, f_\gamma^\mathrm{OoD}\})$ (eq. \ref{eq:acquisition_function})
  \STATE Label $A^s$
  \STATE Add $A^s$ to labeled data $\mathcal{D}_\mathit{train}^{s+1} \gets \mathcal{D}_\mathit{train}^{s} \cup A^s$
  \STATE Remove $A^s$ from pool $\mathcal{D}_\mathit{pool}^{s+1} \gets \mathcal{D}_\mathit{pool}^{s} \setminus A^s$
  \ENDFOR
 \RETURN Optimal model parameters $\alpha, \beta, \gamma$; training dataset $\mathcal{D}_\mathit{train}^S$.
 \end{algorithmic}
 \end{algorithm}
 
\section{Experiments} \label{sec:experiments}
For the empirical validation, we use two publicly available datasets, Panda and MNIST. In the MNIST dataset, we artificially introduce ambiguities and artifacts to demonstrate the functionality of the different model components. The proposed FocAL strategy avoids ambiguities, and artifacts and outperforms other approaches. In the second experiment with the Panda dataset, we apply the model on real-world data. We perform a study about the introduced hyperparameters, analyze the uncertainty estimations of the model and compare different AL methods. compare to the following other AL strategies that were used in the recent literature:
\\
\emph{RA} \cite{gal_deep_2017, carse_active_2019, raczkowski_ara_2019}:
 Random Acquisition (RA) is a simple baseline method that uses a uniform distribution over the images instead of an informativeness measure.
\\
\emph{EN} \cite{gal_deep_2017, zeng_relevance_2018, raczkowski_ara_2019}:
The maximum entropy (EN) is used for acquisition. As entropy is a measure of new information, the most informative images should be obtained.
\\
\emph{BALD} \cite{houlsby_bayesian_2011, gal_deep_2017, carse_active_2019, raczkowski_ara_2019}:
Acquisition with Bayesian Active Learning by Disagreement (BALD). The idea of BALD is to select the images which maximize the mutual information between predictions and model posterior. This is one of the most popular methods adapted in recent literature.
\\
\emph{MS} \cite{gal_deep_2017}:
The Mean Std (MS) measures the uncertainty by the average standard deviation of the predictive distribution. The idea is to acquire images with the least confident predictions.
\\
\emph{EP} \cite{nguyen_epistemic_2019}:
BNN using only the epistemic uncertainty as calculated in equation \ref{eq:bnn_uncertainties}. This method is similar to FocAL, but without weighting the epistemic uncertainty and without the aleatoric uncertainty and OoD scoring.
\\
\emph{FocAL}:
The proposed FocAL method as described in Section \ref{sec:method}.
\subsection{MNIST}
The goal of this experiment is to illustrate the functionality of FocAL in a controlled environment with an intuitive dataset with artificial artifacts and ambiguities.\\
\noindent 
\textbf{Dataset}
The well-known MNIST dataset \cite{deng_mnist_2012} contains 60,000 training and 10,000 test images of handwritten digits with 28x28 greyscale pixels. Of the original training split, we randomly sample 2000 images of which 20 images are initially labeled ($D_\mathit{train}$) while 1980 images remain initially unlabeled ($D_\mathit{pool}$). This relatively low number of images is chosen for better visualization of the data distribution (Fig. \ref{fig:data_dist} and \ref{fig:mnist_acquisition}). In each acquisition step, 10 images are acquired (labeled) until $\mathcal{D}_\mathit{train}$ contains 200 labeled images.
We also sample 200 validation images (from the training split) to use a reasonably small validation set in the context of limited labeled data \cite{oliver_realistic_2018}. For testing, we use the original test split of 10000 images.
Furthermore, we adjust this dataset to mimic the problems in digital pathology that we want to tackle. 
The class imbalance is obtained by reducing the original 10 classes to only 3 classes: Digit '0', digit '1', and 'all other digits'. The classes '0' and '1' represent the malignant classes  (10\% portion of the whole dataset each) while 'all other digits' represent the healthy tissue (80\% portion of the whole dataset).

\begin{figure}[!t]
     \centering
     \includegraphics[width=0.8\linewidth]{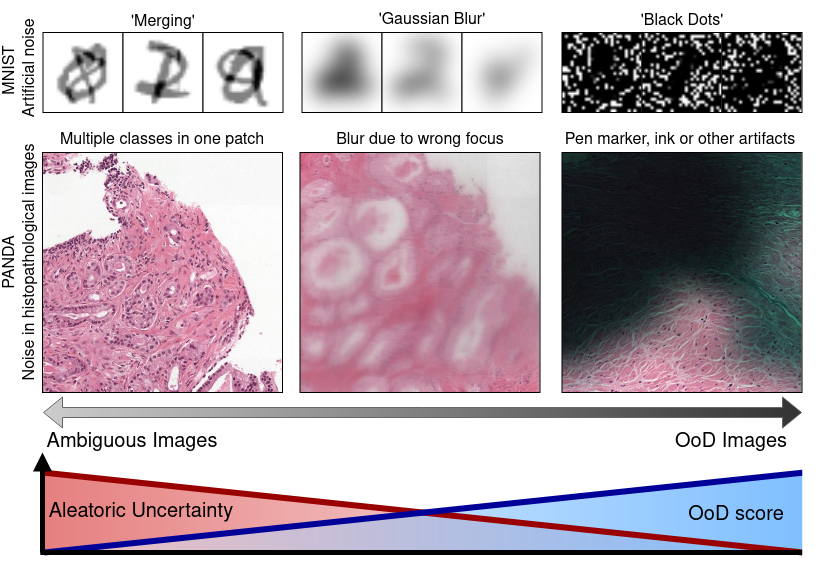}
     \caption{
     Images with ambiguities and artifacts that should be avoided during AL acquisition. The top row shows MNIST images with the artificial noise types 'Merging', 'Gaussian Blur', and 'Black Dots'. They simulate the artifacts and ambiguities encountered in histopathological images (bottom row) in the Panda dataset. The left Panda patch contains two different classes (Gleason Grade 3 and 4), the middle patch is blurry due to wrong microscope focus and the right patch is covered by pen marker, obscuring most tissue parts. Although a clear categorization is difficult, we propose the following scale: The images on the left side show ambiguities but the images are in-distribution because their appearance (color distribution and shapes) is normal. The images on the right side can be considered OoD because the color distribution and shapes substantially differ from the 'normal' images of interest. The blurry images are in between these two extremes as the color distribution and appearance is slightly OoD and they contain ambiguities due to blurry edges and patterns. We will see that with the proposed FocAL method, the shown images are avoided thanks to the aleatoric uncertainty and OoD score.} \label{fig:noisy_image_examples}
\end{figure}

\begin{figure}[!t]
     \includegraphics[trim={1cm 0cm 1.5cm 0cm}, width=0.65\linewidth]{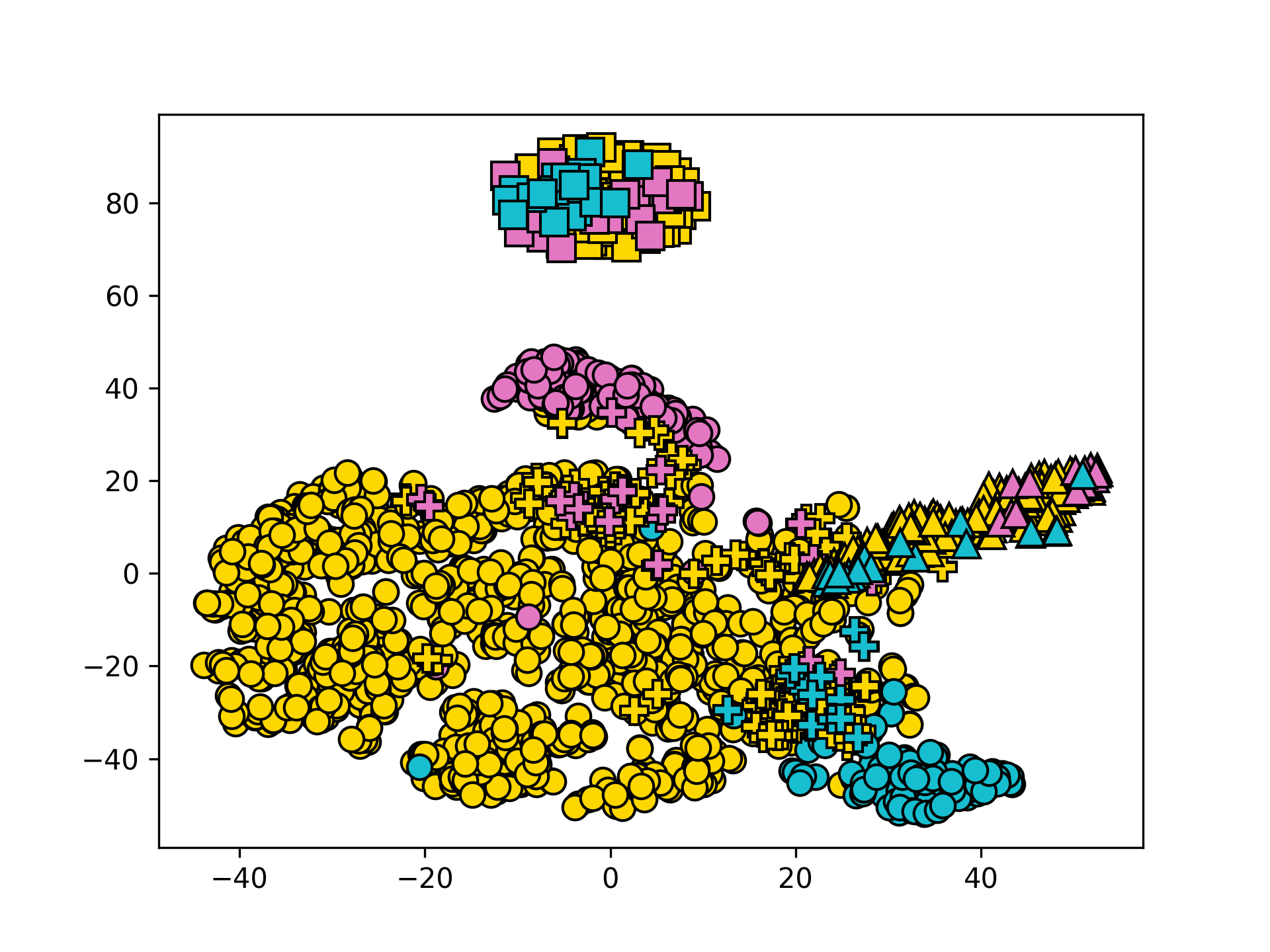}
  \includegraphics[ width=0.18\linewidth]{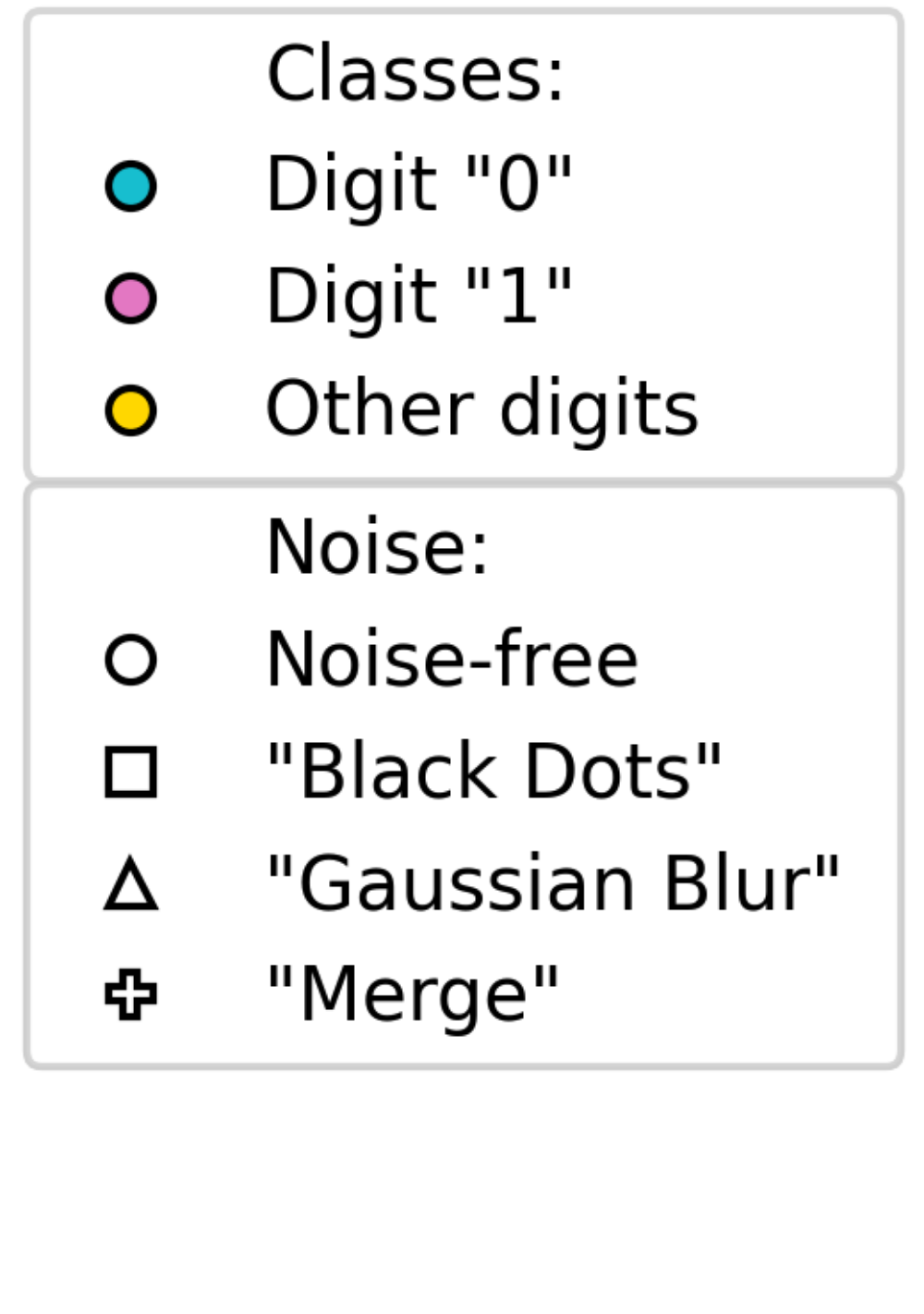}
     \caption{
     Feature distribution of the 2000 images ($X_\mathit{train}$ and $X_\mathit{pool}$) after the last acquisition step with 200 labeled images. Each point represents the feature vector $z$ of an MNIST image, reduced to two dimensions by t-SNE. The distribution supports our categorization of artifacts and ambiguities  (Fig. \ref{fig:noisy_image_examples}). The images with 'black dots' (depicted as squares) are OoD while the 'merged' images are ambiguous and therefore close to the class boundaries. Blurred images show both characteristics (OoD and ambiguities) as some images are far away from the distribution of interest, while others lie close to the class boundaries.
     } \label{fig:data_dist}
\end{figure}

\begin{figure*}[!t]
     \centering
     \subfloat[FocAL epistemic unc.\label{fig:f_a}]{
         \centering
         \includegraphics[trim={1cm 0cm 1cm 0cm}, width=0.33\linewidth]{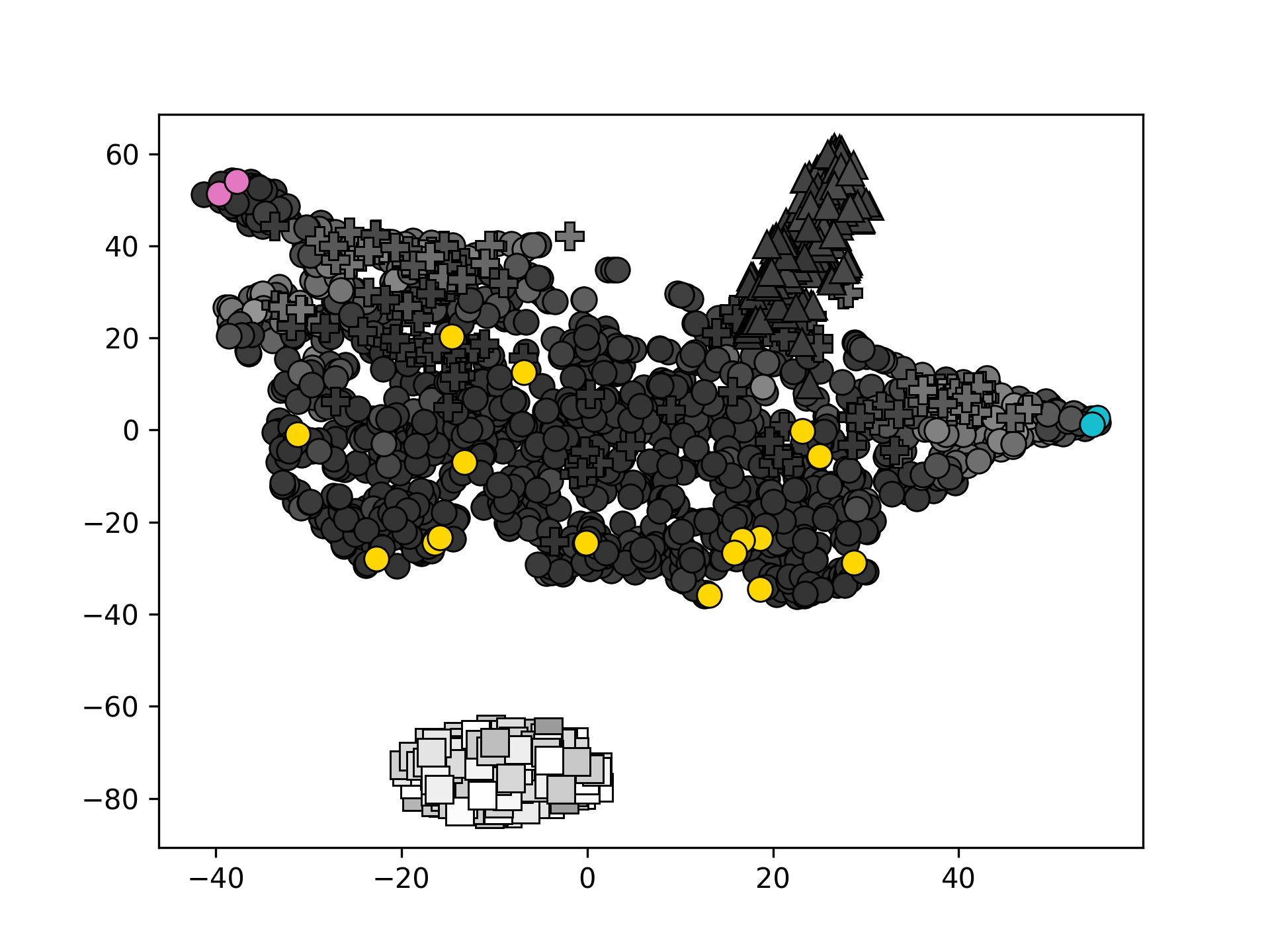}
     }
    \subfloat[FocAL aleatoric unc.\label{fig:f_b}]{
         \centering
         \includegraphics[trim={1cm 0cm 1cm 0cm},width=0.33\linewidth]{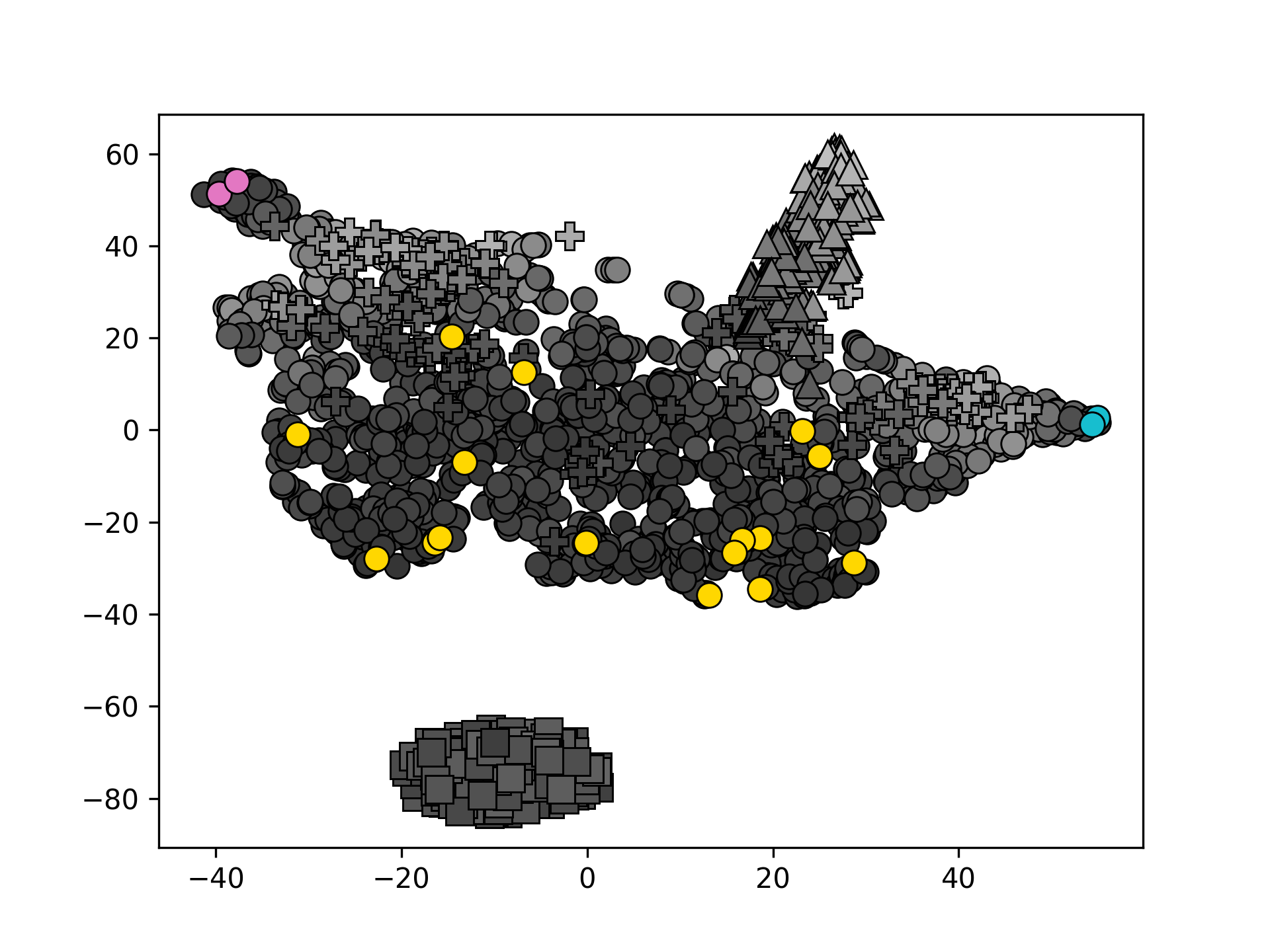}
     }
    \subfloat[FocAL OoD score \label{fig:f_c}]{
         \centering
         \includegraphics[trim={1cm 0cm 1cm 0cm},width=0.33\linewidth]{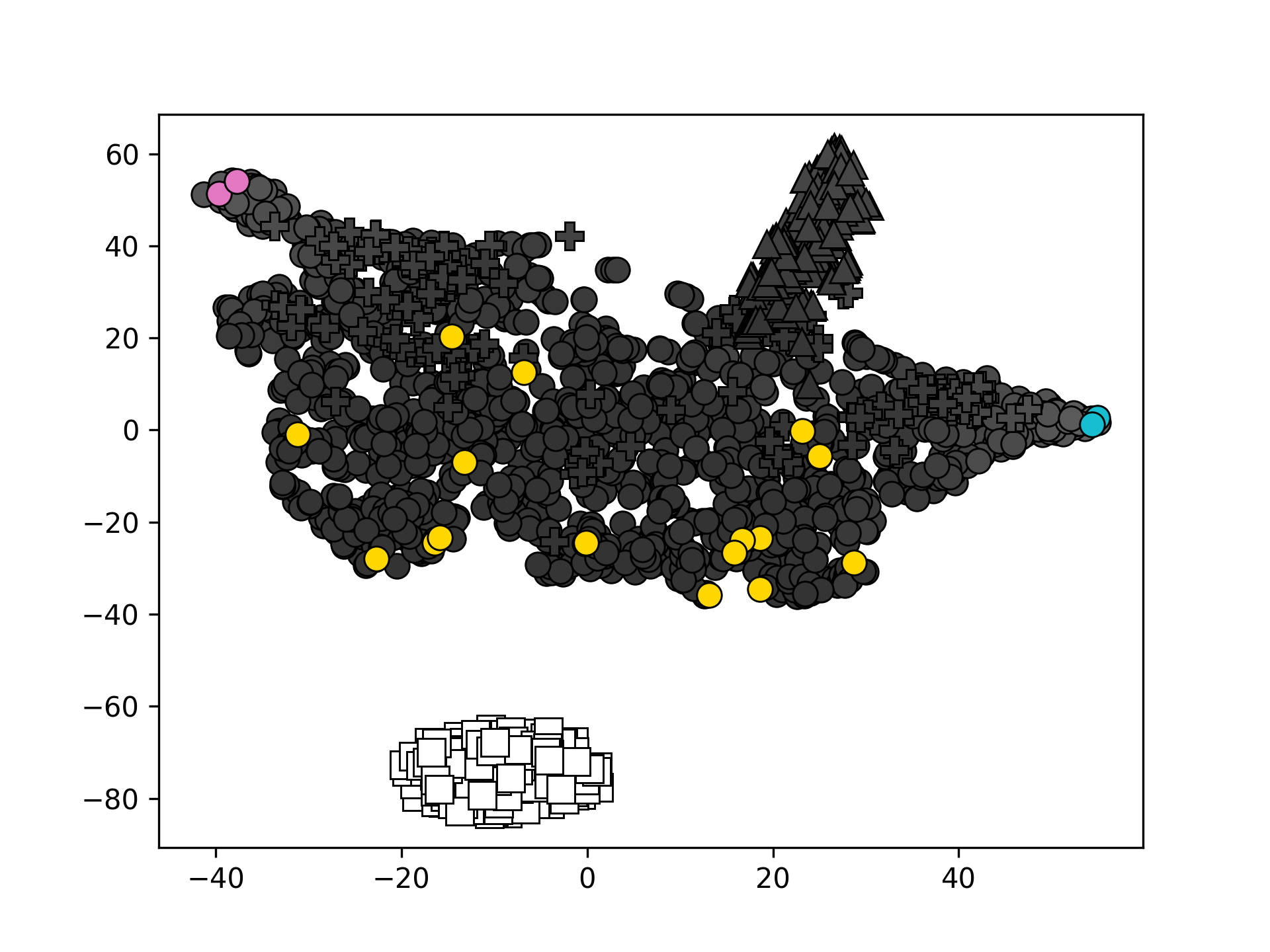}
     } \\
     \subfloat[FocAL acquisition score\label{fig:f_d}]{
         \centering
         \includegraphics[trim={1cm 0cm 1cm 0cm},width=0.33\linewidth]{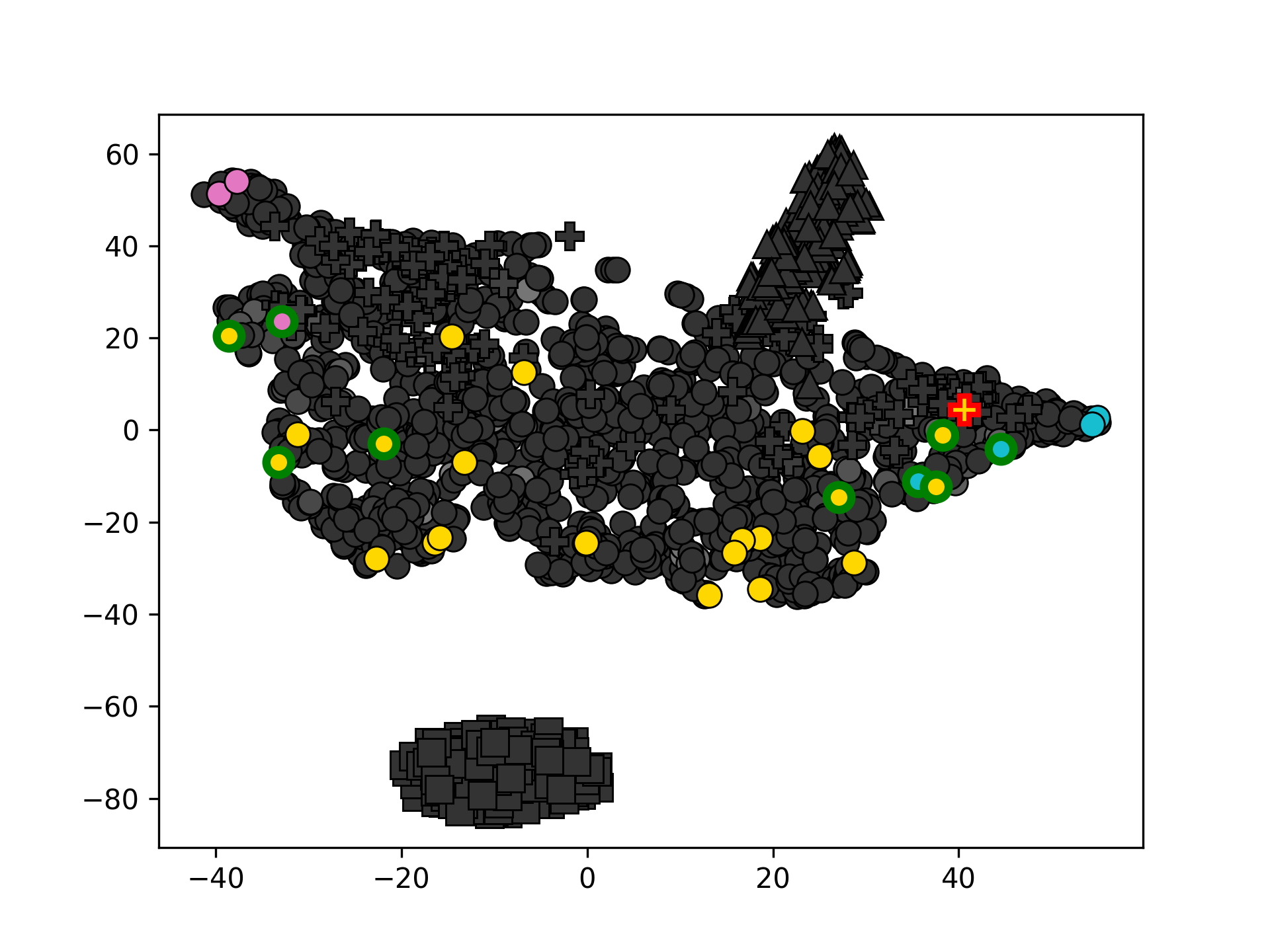}
     } 
    \subfloat[EN acquisition score\label{fig:f_e}]{
         \centering
         \includegraphics[trim={1cm 0cm 1cm 0cm},width=0.33\linewidth]{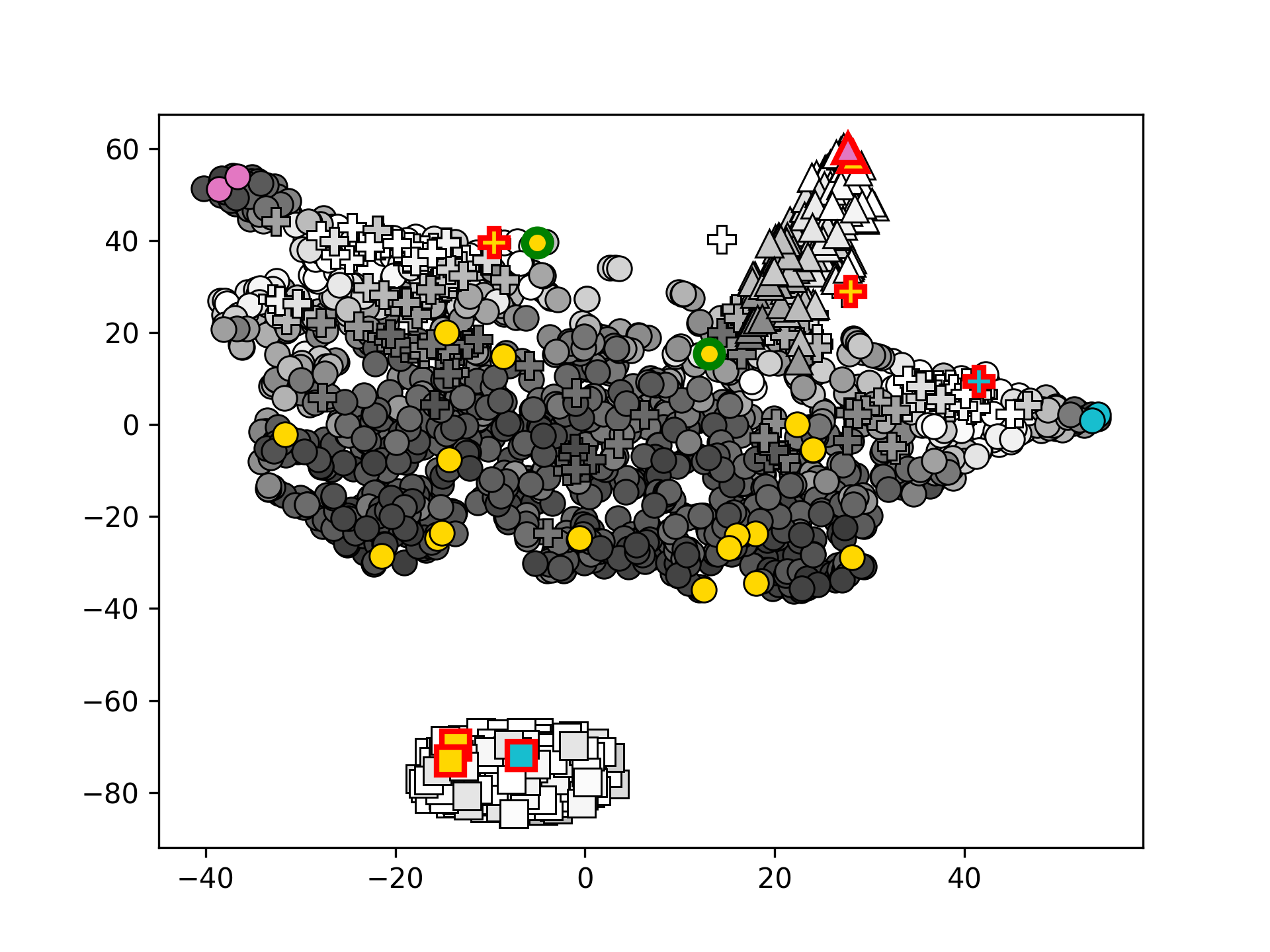}
     }
    \subfloat[BALD acquisition score\label{fig:f_f}]{
         \centering
         \includegraphics[trim={1cm 0cm 1cm 0cm},width=0.33\linewidth]{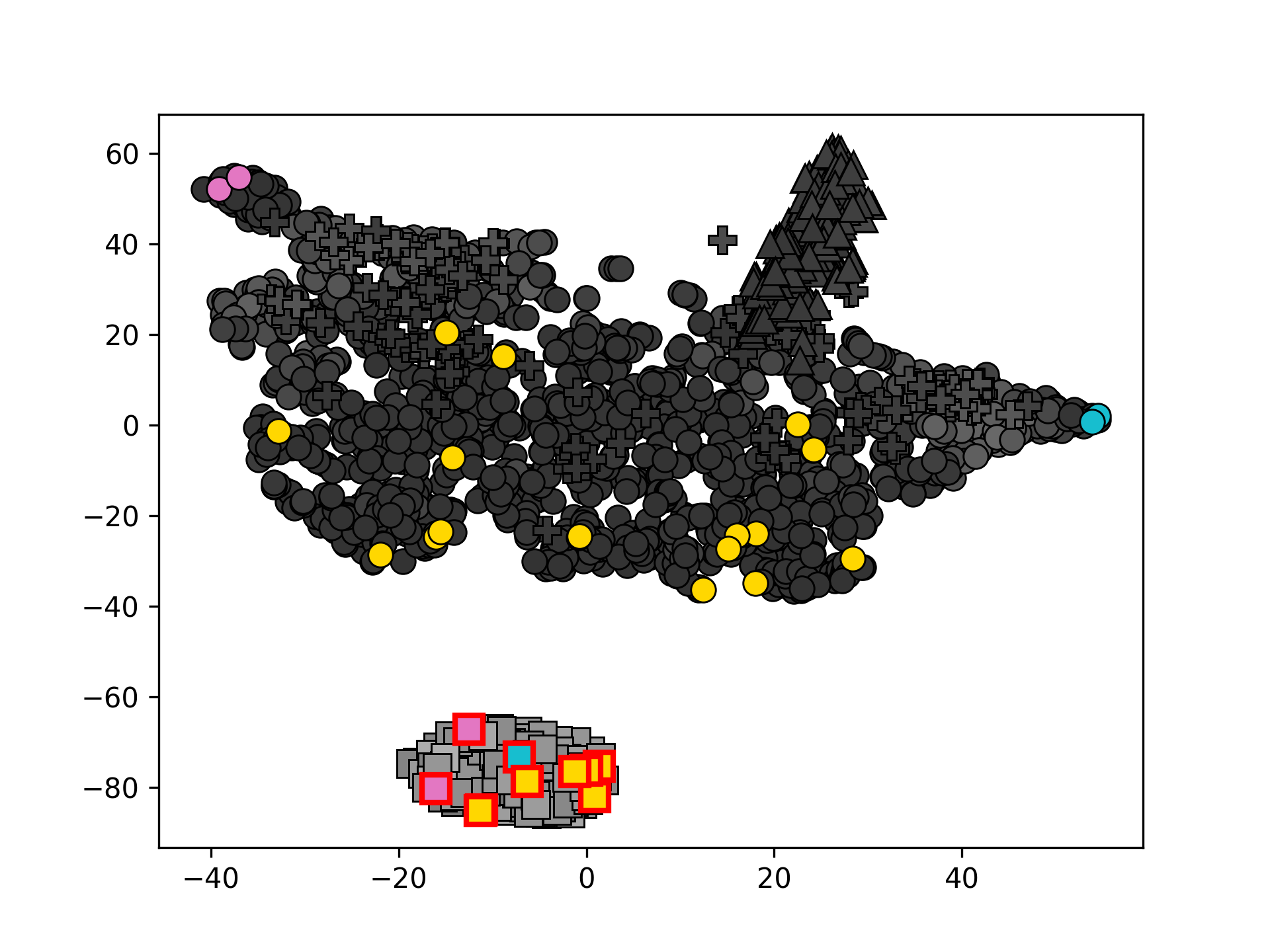}
     }\\
     \subfloat{         \includegraphics[width=0.5\linewidth]{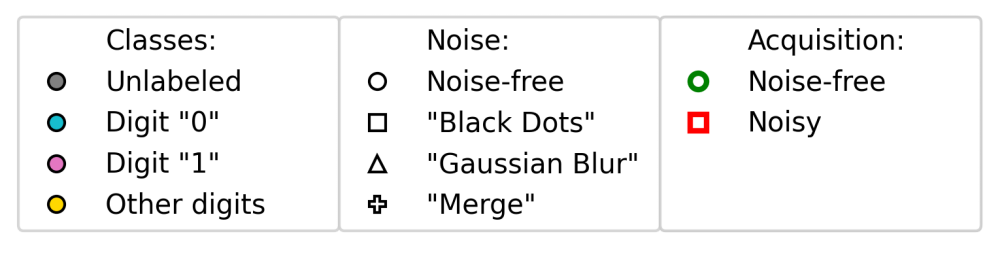}}
        \caption{
        Feature distribution, uncertainty, and acquisition scores for the first acquisition step (best viewed with zoom), similar to Figure \ref{fig:data_dist}. Labeled images are dots filled with turquoise (Digit "0"), pink (Digit "1"), or yellow (other digits). Unlabeled images are dots with greyscale color, representing the uncertainty or acquisition score (the higher the brighter). The datapoints with a green edge represent noise-free images (good for training) and datapoints with a red edge represent images with artifacts, blur or ambiguities. 
        For the proposed FocAL method, the epistemic uncertainty (a) measures the image informativeness, but it is easily distracted by artifacts and ambiguities. These uninformative images can be captured by a high aleatoric uncertainty (b) or a high OoD score (c). Therefore, in the final FocAL acquisition (d), 9 noise-free images are acquired (and only 1 ambiguous image).
        The competing methods EN (e) and BALD (f) in comparison acquire almost only images with artifacts or ambiguities in this step which add less information to the training.}
        \label{fig:mnist_acquisition}
\end{figure*}

\noindent 
\textbf{Artifacts and Ambuiguities}
The artificial artifacts and ambiguities are obtained by adding perturbations to the input images, as depicted in Figure \ref{fig:noisy_image_examples}. We use three different perturbations that mimic artifacts and ambiguities in histopathological images: \emph{Black dots} are randomly added to 75\% of the total image pixels by setting the greyscale value to $0$. This simulates pen marker or ink in histopathological images that can cover large parts of image patches. \emph{Gaussian blur} filter with a standard deviation of $\sigma=4$ is used to simulate the blur caused by wrong focus. \emph{Merging} by randomly blending one image with another image of a different class together leads to ambiguous images with two different plausible labels (while maintaining the original one-hot encoded label). This simulates ambiguities by the presence of two cancerous classes in one image patch or edge cases with unclear ground truth label. Each of the three perturbations is applied to a total of 200 unlabeled images.

\noindent 
\textbf{Implementation Details}
As the images of MNIST are very small, we use a simple feature extractor consisting of one convolutional layer with 4 filters (stride 3x3), max pooling (stride 2x2), and one fully connected layer with 128 units. The BNN consists of two fully connected layers with 128 units each and a final softmax layer with three output units, corresponding to the three classes. 
We use the cross-entropy loss and the Adam optimizer \cite{kingma_adam_2015} for 1000 epochs before each acquisition step. The learning rate is set to 0.0001 and multiplied by 0.5 if the validation accuracy does not increase for 50 epochs. The combination of a high number of epochs and learning rate reduction assures complete convergence at each acquisition step. We experimentally set the weight factors to $\lambda_{al} = 0.5$ and $\lambda_{ood} = 2.0$ since it showed the best results (tested: $0.5$, $1.0$, and $2.0$ for each hyperparameter). Note, that an extensive ablation study of these hyperparameters is included in Section \ref{subsec:exp_panda} for the Panda dataset.

\noindent 
\textbf{Data Distribution}
First, we empirically analyze the data distribution with respect to the artificial artifacts and ambiguities. For this purpose, we plot the feature vectors $z$ after the complete AL process using the FocAL method. We reduce the features to a two-dimensional distribution with t-SNE and depict the data in Figure \ref{fig:data_dist}.
The distribution empirically reflects the categorization shown in Fig. \ref{fig:noisy_image_examples}. The images with 'black dots' are OoD because they are far away from the data distribution of interest. The images with 'Gaussian blur' are partially OoD and the 'merged' images are completely in-distribution. Similarly, ambiguities can be identified. The 'merged' images and a part of the images with 'Gaussian blur' are ambiguous and therefore close to the class boundaries. The images with 'Black dots' are not ambiguous. Apart from this data-related observation, the figure shows that the model learns to separate the classes during the active learning procedure. This class-separation is a necessary step for a good final classification.

\noindent 
\textbf{Acquisition}
Figure \ref{fig:mnist_acquisition} shows the acquisition behavior of FocAL and competing methods. 
It confirms that the FocAL model components work as expected in practice.
The weighted epistemic uncertainty of the FocAL method (\ref{fig:f_a}) is high (bright greyscale color) for images at the class boundaries, but also for noisy images (especially of the noise type 'Black dots' and 'Merging'). This means that it is a good measure of informativeness, but 'distracted' easily by artifacts and ambiguities. The aleatoric uncertainty (\ref{fig:f_b}) captures images with 'Merging' and 'Gaussian blur' while the OoD score (\ref{fig:f_c}) highlights the images with 'Black dots'. These images with ambiguities and artifacts are avoided during acquisition. As a result, the FocAL method (\ref{fig:f_d}) acquires only 1 ambiguous image while 9 acquired images are informative and contain several images of the minority classes '0' and '1'. 
Furthermore, the acquisition analysis highlights the problems of existing AL methods.
EN in Figure \ref{fig:f_e} and BALD in Figure \ref{fig:f_f} are highly 'distracted' by images with ambiguities and artifacts and do not acquire any informative data in this step. Ambiguous images are close to class boundaries while OoD images have a 'novel' image content due to their different appearance. Therefore, the acquisition scores of other AL methods for these images are usually high. Although here we depict only the data distribution of features from the first acquisition step, these observations are representative for all further acquisition steps as well. 

\begin{figure}[!t]
     \centering
     \subfloat[Acquired images with artifacts or ambiguities\label{fig:mnist_results_a}]{
         \centering
         \includegraphics[width=0.45\linewidth]{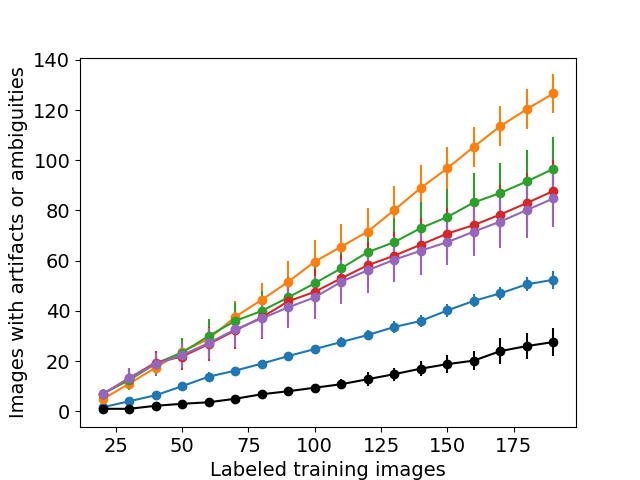}
     }
    \subfloat[Test accuracy \label{fig:mnist_results_b}]{
         \centering
         \includegraphics[width=0.45\linewidth]{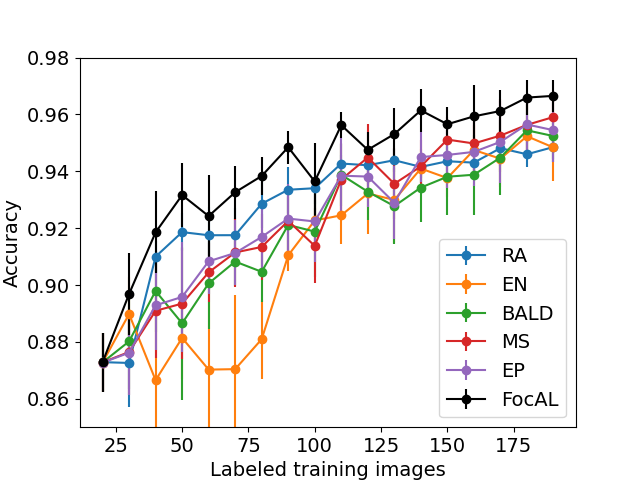}
     }\\
        \caption{Results of the MNIST Experiments with mean and standard error of five independent runs. In \ref{fig:mnist_results_a} the total acquired noisy images are plotted and in \ref{fig:mnist_results_b} and the accuracy. The proposed FocAL algorithm effectively avoids acquiring images with ambiguities and artifacts and shows the strongest performance. }
        \label{fig:mnist_results}
\end{figure}

\noindent 
\textbf{Model Comparison}
Figure \ref{fig:mnist_results_a} confirms the previous observation (Fig. \ref{fig:f_e} and \ref{fig:f_f}) that other methods acquire many images with artifacts and ambiguities, even more than the model with random acquisition (RA). Figure \ref{fig:mnist_results_b} shows that the acquisition of many images with ambiguities and artifacts harms the test performance, as the model EN acquires the most images with ambiguities and artifacts and shows the worst performance. The other AL methods also acquire a substantial amount of images with ambiguities and artifacts and their performance is on par or even below the baseline model with random acquisition RA. Note that in other studies AL methods outperform random acquisition but many are conducted on clean, highly curated datasets. This often does not apply to real-world data (like histopathological images). The proposed FocAL method effectively avoids acquiring uninformative images which leads to the overall best performance. 

To compare to the supervised baseline, we try two different settings: one trained with all 2000 images (including the 600 images with artifacts and ambiguities) and another one with only the 1400 images without perturbations. Again, we report the average results over 5 independent runs. The supervised model trained without artifacts and ambiguities performed substantially better (accuracy 0.965; mean f1 score: 0.940) than the model trained with all 2000 images (accuracy 0.939; mean f1 score: 0.901). This confirms our hypothesis that avoiding artifacts and ambiguities is essential for a good performance. We observe that the FocAL method reaches the supervised performance (with all 2000 images) with only 90 labeled training images ($4.5\%$ of all images) and even outperforms this supervised performance due to the successful avoidance of perturbed images. Therefore, FocAL not only alleviates the labeling process - it can  further save resources usually necessary for data curation. The performance of the supervised model trained on the highly curated dataset with 1400 images is reached by FocAL with 190 labeled images ($9.5\%$ of all training data). 

\subsection{Panda}  \label{subsec:exp_panda}
The Panda dataset is a large open dataset for the classification of prostate cancer. We use this dataset for hyperparameter tuning, analysis of the different uncertainty measures, and finally for a comparison of the different AL methods.

\noindent 
\textbf{Dataset}
The Panda dataset \cite{bulten_artificial_2022} consists of 10,616 WSIs and was presented at the MICCAI 2020 conference as a Kaggle challenge. 
Two institutes participated in labeling the WSIs. The images from Radboud University Medical Center come with detailed label masks of all tissue parts in the Gleason Grading (GG) scheme. The classes are 'Non-cancerous' (NC), 'Gleason 3', 'Gleason 4', and 'Gleason 5' depending on the architectural growth patterns of the tumor. The second institute, the Karolinska Institute, only assigned binary (cancer vs. healthy) labels and we therefore disregard their images for our experiments. After sorting out corrupted images, a total of 5058 WSIs are left of which we used 1000 WSIs for testing and 30 WSIs for validation that were randomly chosen.
The WSIs were divided into 50\% overlapping 512x512 patches. To use a multi-scale feature extractor (see Implementation paragraph below), we determine the class of each patch by its center segment (256x256 square), depending on the majority class of the pixels if at least 5\% of the pixels are annotated as cancerous. If less than 5\% of the pixels are annotated as cancerous, the patch is assigned the non-cancerous label. If a patch contains more than $95\%$ of background (according to the annotation mask), it is disregarded. Therefore, the dataset is already curated because several artifacts are already excluded in this step. Note that in other experiments, where the dataset curation is more difficult, the advantage of FocAL might be even bigger.

For the empirical validation, we design two different experiments, \emph{small-Panda} for studying the hyperparameter setting due to computational constraints and \emph{big-Panda} with all available images for the final experiments.
In the small-Panda experiment, we take a subset of 200 WSIs for training. For the AL start, we randomly choose 20 WSIs with 5 labeled patches each, resulting in 100 labeled patches in total (equally  distributed over the  classes with 25 patches per class). We perform 16 acquisition steps of 50 patches each. 
The big-Panda setup includes all available 4028 WSIs from the Radboud center for training. To reach a competitive performance with limited computational resources, we start with 400 randomly extracted patches of 100 WSIs (100 from each class). In each acquisition step, we acquire 400 more patches. Validation and test sets are equal for both experiments, see above. The common metric to measure the classification of prostate cancer is Cohen's quadratic kappa which measures the similarity of the ground truth and the predictions, taking the class order into account (misclassifying Gleason 5 as Gleason 4 has less impact than misclassifying Gleason 5 as NC).

\noindent 
\textbf{Implementation}
For the feature extraction, we use the EfficientNetB3 model \cite{tan_efficientnet_2019} as a CNN backbone in a multi-scale architecture. 
Remember that the patch classes were assigned based on the center 256x256 square of each patch with resolution 512x512 (see dataset paragraph above).
The center square (256x256) is cropped and fed into the CNN. At the same time, the complete 512x512 patch is resized to 256x256 and fed into a parallel CNN. The two feature vectors (of 1536 dimensions each) are concatenated, followed by a dropout layer and a fully connected layer with 128 units. This approach has the following two advantages. First, the WSI can be segmented with a high level of detail because the classification is performed for relatively small patch centers of 256x256. Second, the context (surrounding tissue) can still be taken into account by the model. 

The complete feature extractor has fewer than 22 million trainable parameters (for comparison, a ResNet50 \cite{he_deep_2016} has over 23 million parameters). The BNN consisted, as in the MNIST experiment, of two fully connected layers with 128 units and a final softmax layer with one output unit per class. 
We train the model with the Adam optimizer \cite{kingma_adam_2015} for 200 epochs for each acquisition step. The learning rate is set to $1\mathrm{e}{-4}$ for the first 100 epochs, then reduced to $1\mathrm{e}{-5}$ for the other 100 epochs.

\begin{figure}[!t]
\centering
\includegraphics[width=0.6\linewidth]{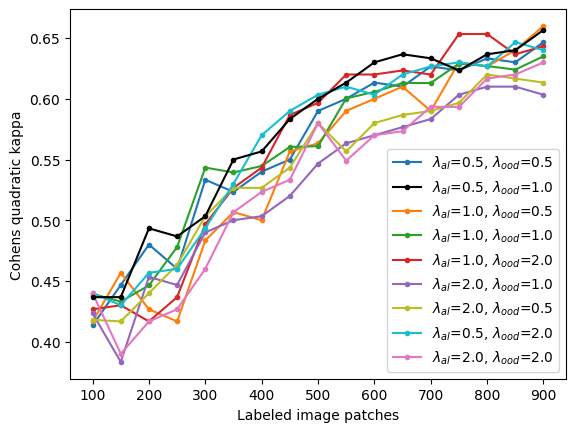}

\caption{Hyperparameter Tuning of the FocAL method. The two hyperparameters $\lambda_{al}$ and $\lambda_{ood}$ are analyzed which weight the aleatoric uncertainty and OoD detection, respectively. The best performing model with $\lambda_{al}= 0.5, \ \lambda_{ood}= 1.0$ is used for the final experiments. If the weight $\lambda_{al}$ is chosen too high, the performance of the model drops because the avoidance of ambiguities is given too much importance.}
\label{fig:ablation_studies}
\end{figure}

\begin{figure*}[!t]
     \centering
     \subfloat[WSI \label{fig:wsi_a}]{
         \centering
         \fbox{\includegraphics[trim={0cm 0cm 0cm 0cm}, width=0.1\linewidth]{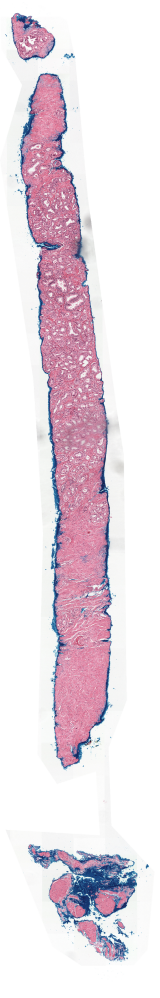}}
     }
    \subfloat[Ground tr. \label{fig:wsi_b}]{
         \centering
         \fbox{\includegraphics[trim={0cm 0cm 0cm 0cm}, width=0.1\linewidth]{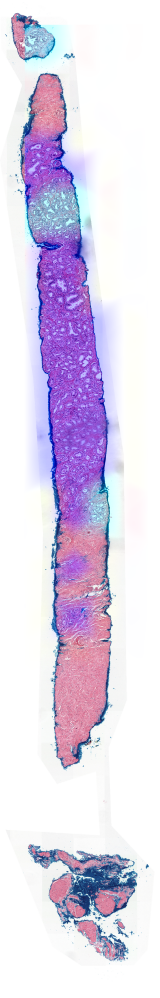}}
     }
     \subfloat[Prediction \label{fig:wsi_c}]{
         \centering
         \fbox{\includegraphics[trim={0cm 0cm 0cm 0cm}, width=0.1\linewidth]{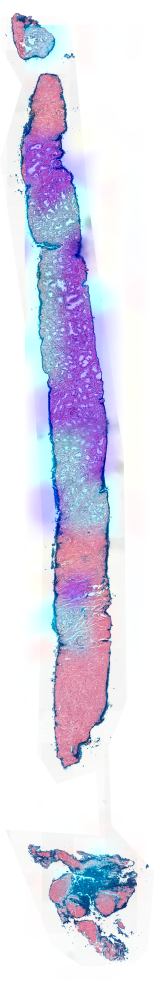}}
     } 
    \subfloat[Epist. unc. \label{fig:wsi_d}]{
         \centering
         \fbox{\includegraphics[trim={0cm 0cm 0cm 0cm}, width=0.1\linewidth]{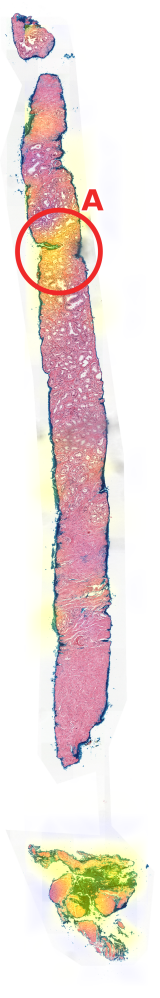}}
     }
    \subfloat[Aleat. unc. \label{fig:wsi_e}]{
         \centering
         \fbox{\includegraphics[trim={0cm 0cm 0cm 0cm}, width=0.1\linewidth]{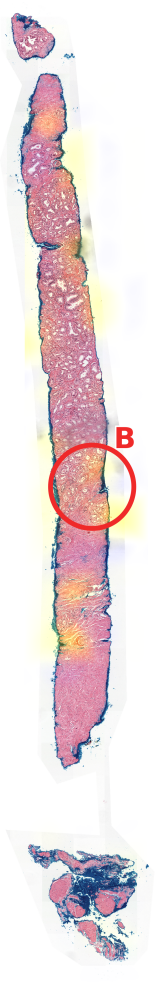}}
     }
     \subfloat[OoD Score \label{fig:wsi_f}]{
         \centering
         \fbox{\includegraphics[trim={0cm 0cm 0cm 0cm}, width=0.1\linewidth]{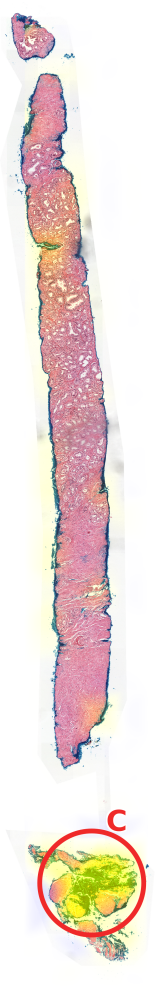}}
     }
     \subfloat[Acquis. Sc. \label{fig:wsi_g}]{
         \centering
         \fbox{\includegraphics[trim={0cm 0cm 0cm 0cm}, width=0.1\linewidth]{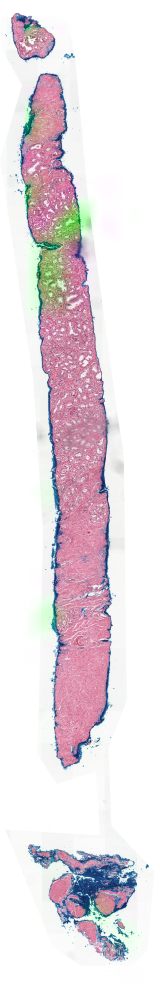}}
     }
    \subfloat{
         \centering \includegraphics[trim={0cm 0cm 0cm 0cm}, width=0.1\linewidth]{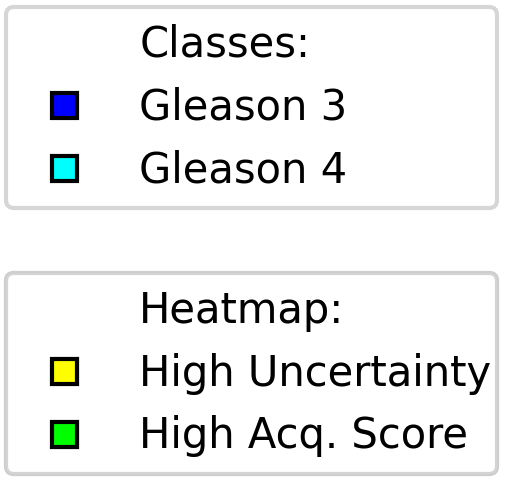}
     }
        \caption{Visualization of predictions and uncertainties for a test slide at the first acquisition step of the Panda dataset (after training on the initial 400 patches). It was split into 120 overlapping patches and all heatmaps were produced by cubic interpolation of the patch center predictions. Uncolored parts correspond to non-cancerous tissue (in \ref{fig:wsi_b}, \ref{fig:wsi_c}) and areas without (or with low) uncertainties (in \ref{fig:wsi_d} - \ref{fig:wsi_g}). The red circle A in Fig. \ref{fig:wsi_d} marks an area with high epistemic uncertainty. As the aleatoric uncertainty and OoD score are low in this area, this results in a high acquisition score (green in Fig. \ref{fig:wsi_g}). The area of circle A is informative. The red circle B in Fig. \ref{fig:wsi_e} shows an area with high aleatoric uncertainty due to slight blur and ambiguities. Therefore, the final acquisition score is low in this area. The red circle C in Fig. \ref{fig:wsi_f} shows an area with artifacts (blue ink) that results in a high OoD score. Although the epistemic uncertainty is high in this area, the acquisition of these non-cancerous and uninformative patches is avoided.}
        \label{fig:uncertainty_wsi}
\end{figure*}

\noindent 
\textbf{Hyperparameter Tuning}
First, we perform experiments regarding the newly introduced hyperparameters. We analyze $\lambda_{al}$ and $\lambda_{ood}$ of the FocAL acquisition function (eq. \ref{eq:acquisition_function}) that weight the importance of avoiding ambiguities and OoD images, respectively. For this purpose, we use the small-Panda setup (as described in the dataset paragraph above). Figure \ref{fig:ablation_studies} shows the results for different hyperparameter settings. If the factors are too high, the model performance decreases, as the worst performances are given for the models with $(\lambda_{al}=2.0, \ \lambda_{ood}= 1.0)$ and $(\lambda_{al}=2.0, \ \lambda_{ood}= 2.0)$. We assume that in this case, the model focuses too much on the avoidance of ambiguities such that the novelty (measured by the epistemic uncertainty) is not given enough importance. Especially a high $\lambda_{al}$ can harm the performance, as this measure sometimes mistakenly shows high values for informative images at the class boundary. 
We choose the model with $\lambda_{al}=0.5, \ \lambda_{ood}=1.0$ for the final experiments because it is the overall best-performing model. Note that a comparable performance was obtained for the models with $(\lambda_{al}=0.5, \ \lambda_{ood}=2.0)$ and $(\lambda_{al}=1.0, \ \lambda_{ood}=2.0)$. Overall the performance is robust for all models with $\lambda_{al}<2$.

\noindent 
\textbf{Uncertainty Estimations}
Figure \ref{fig:uncertainty_wsi} illustrates the uncertainty estimates in the first acquisition step of the FocAL method. It shows that each component of the acquisition function works as expected.
The area with a high acquisition score (\ref{fig:wsi_g}) is based on a high epistemic uncertainty in the circle A in  Figure \ref{fig:wsi_d} and a low aleatoric uncertainty and OoD score in Figures \ref{fig:wsi_e} and \ref{fig:wsi_f}, respectively. Indeed, the area contains cancerous tissue and the model shows some misclassifications here (Gleason 4 instead of Gleason 3), as is clear by comparing Figures \ref{fig:wsi_b} and \ref{fig:wsi_c}. Therefore, labeling these patches can improve the overall model performance. Other parts of the image show a high aleatoric uncertainty, for example, circle B in Figure \ref{fig:wsi_e}. This indicates ambiguous patches and therefore, these images are avoided. The acquisition score in this area is low. In circle C in Figure \ref{fig:wsi_f} we see a region containing mainly artifacts of dark ink. These artifacts are detected by the OoD detection and therefore also avoided in the final acquisition, although the epistemic uncertainty is high in this region.
As the images are produced at an early stage of the active learning process, the model's predictions (\ref{fig:wsi_c}) are not yet accurate which is also reflected by high uncertainty values (\ref{fig:wsi_d} - \ref{fig:wsi_f}). 

\begin{figure*}[!t]
     \centering
     {%
     \setlength{\fboxsep}{0pt}%
     \raisebox{\dimexpr 1.5cm-\height}{EN \hspace{0.4cm} }
     \subfloat[\textbf{{NC}} \label{fig:ena}]{
         \centering
         \fbox{\includegraphics[width=0.13\linewidth]{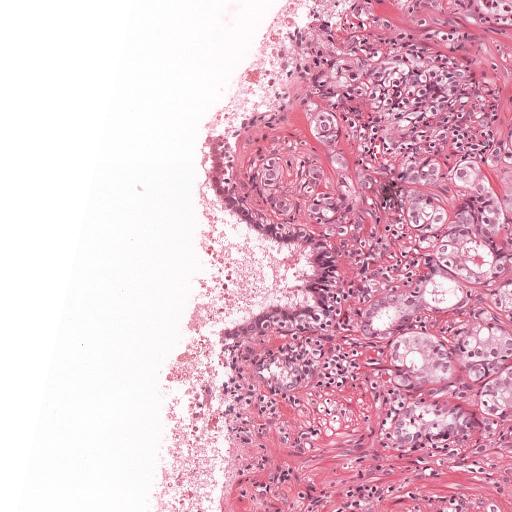}}
     }
     \subfloat[\textbf{\underline{Gleason 4}}\label{fig:enb}]{
         \centering
         \fbox{\includegraphics[width=0.13\linewidth]{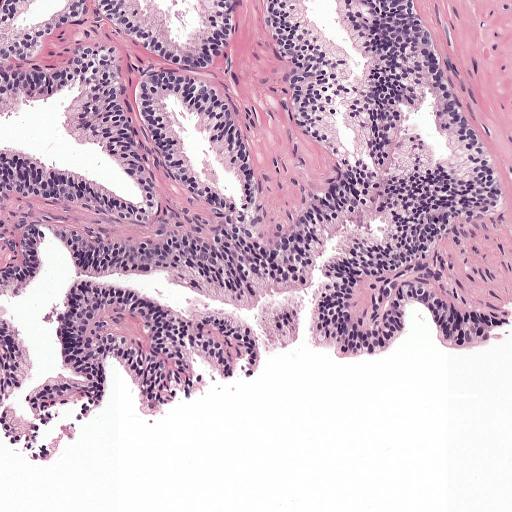}}
     }
     \subfloat[\textbf{\underline{Gleason 4}}\label{fig:enc}]{
         \centering
         \fbox{\includegraphics[width=0.13\linewidth]{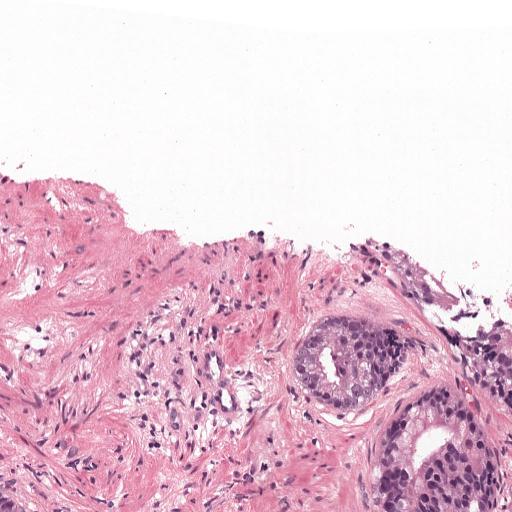}}
     }
    \subfloat[\textbf{NC} \label{fig:end}]{
         \centering
         \fbox{\includegraphics[width=0.13\linewidth]{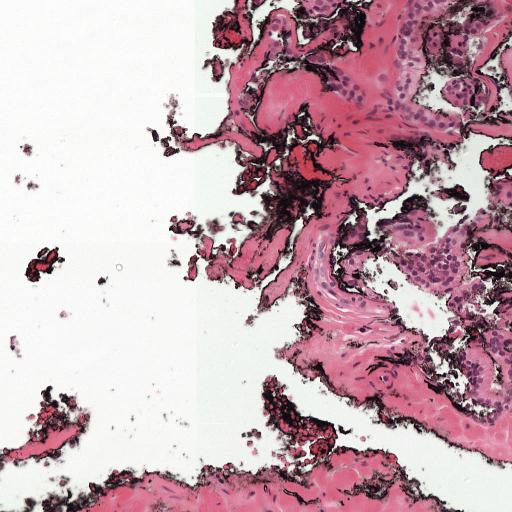}}
     }
    \subfloat[\textbf{\underline{Gleason 3}}\label{fig:ene}]{
         \centering
         \fbox{\includegraphics[width=0.13\linewidth]{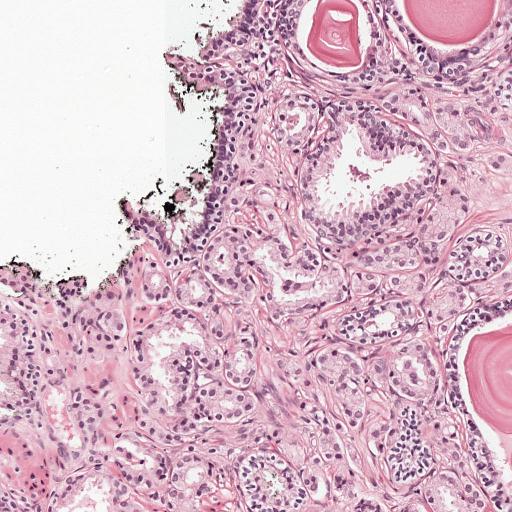}}
     }\\
     \raisebox{\dimexpr 1.5cm-\height}{BALD }
    \subfloat[\textbf{NC} \label{fig:bf}]{
         \centering
         \fbox{\includegraphics[width=0.13\linewidth]{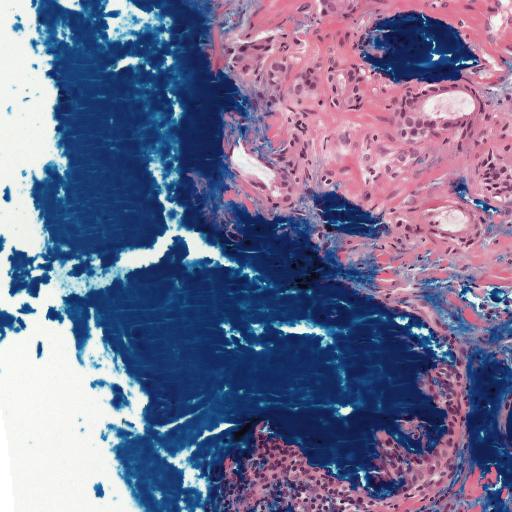}}
     }
    \subfloat[\textbf{NC} \label{fig:bg}]{
         \centering
         \fbox{\includegraphics[width=0.13\linewidth]{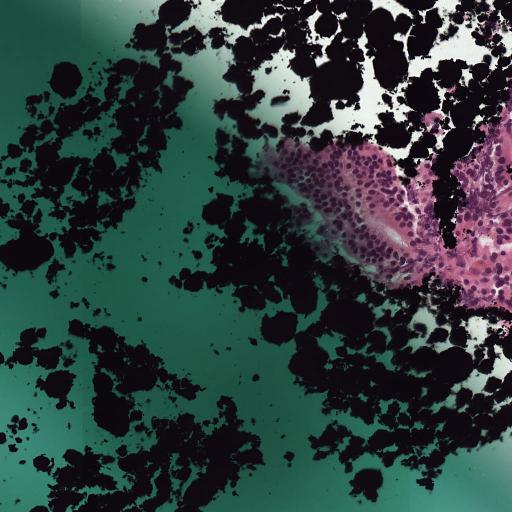}}
    }    
    \subfloat[\textbf{NC}\label{fig:bh}]{
         \centering
         \fbox{\includegraphics[width=0.13\linewidth]{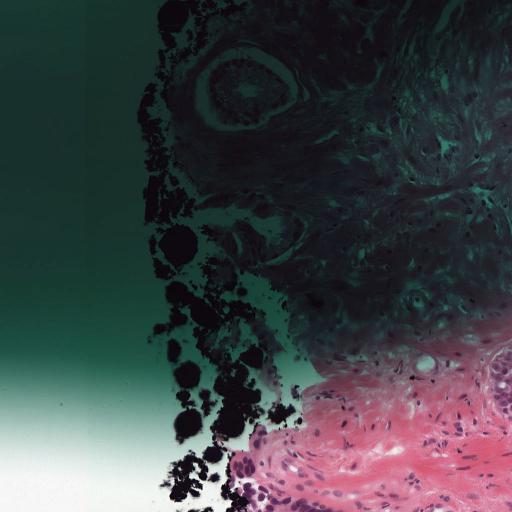}}
     }
    \subfloat[\textbf{NC}\label{fig:bi}]{
         \centering
         \fbox{\includegraphics[width=0.13\linewidth]{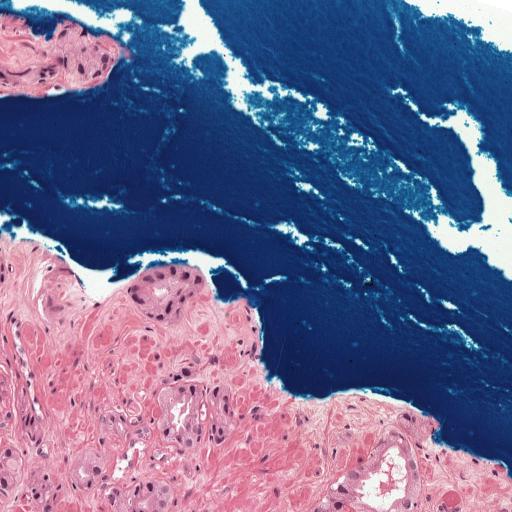}}
     }        
    \subfloat[\textbf{NC}\label{fig:bj}]{
         \centering
         \fbox{\includegraphics[width=0.13\linewidth]{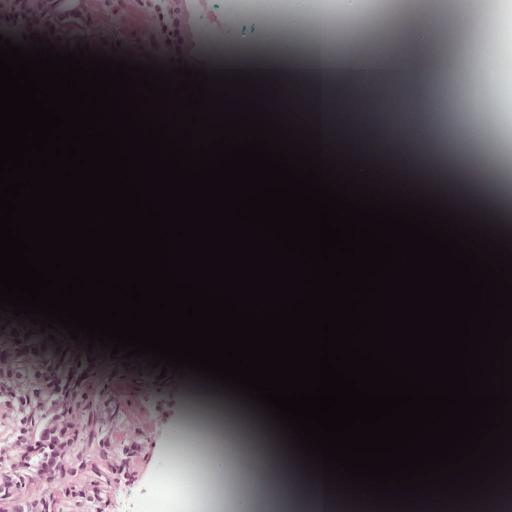}}
    }
    \\
    \raisebox{\dimexpr 1.5cm-\height}{FocAL}
    \subfloat[\textbf{\underline{Gleason 4}}\label{fig:fk}]{
         \centering
         \fbox{\includegraphics[width=0.13\linewidth]{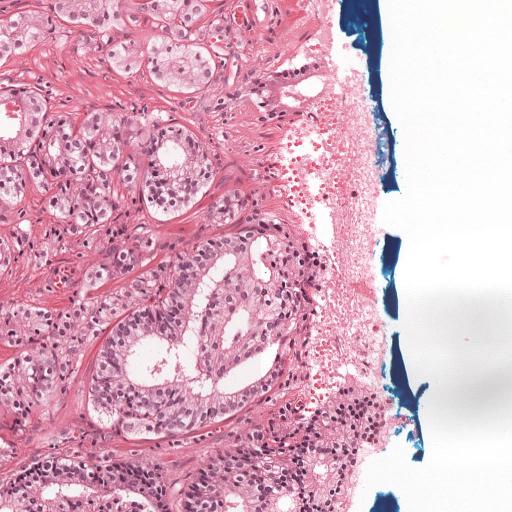}}
     }    
    \subfloat[\textbf{\underline{Gleason 4}}\label{fig:fl}]{
         \centering
         \fbox{\includegraphics[width=0.13\linewidth]{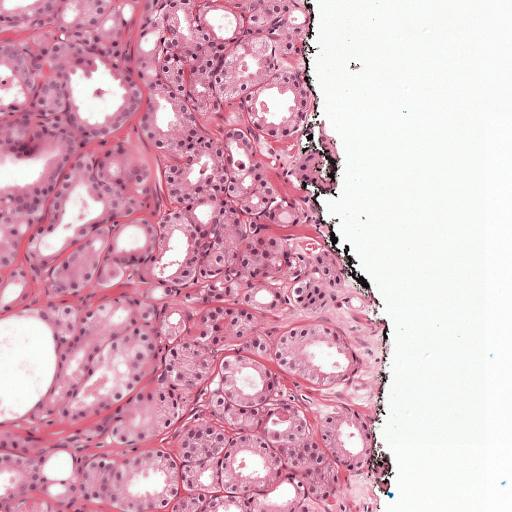}}
     }     
    \subfloat[\textbf{\underline{Gleason 3}}\label{fig:fm}]{
         \centering
         \fbox{\includegraphics[width=0.13\linewidth]{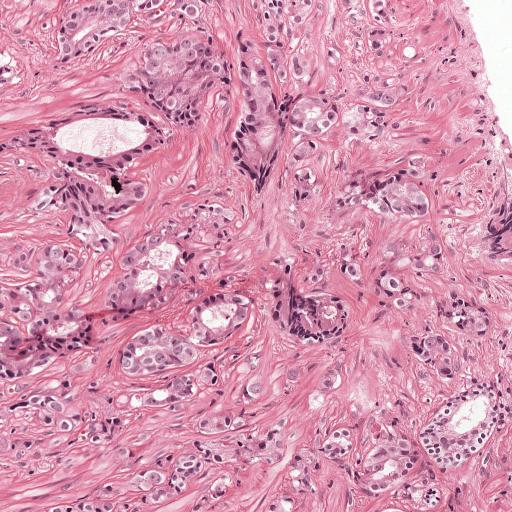}}
     }
     \subfloat[\textbf{\underline{Gleason 4}}\label{fig:fn}]{
         \centering
         \fbox{\includegraphics[width=0.13\linewidth]{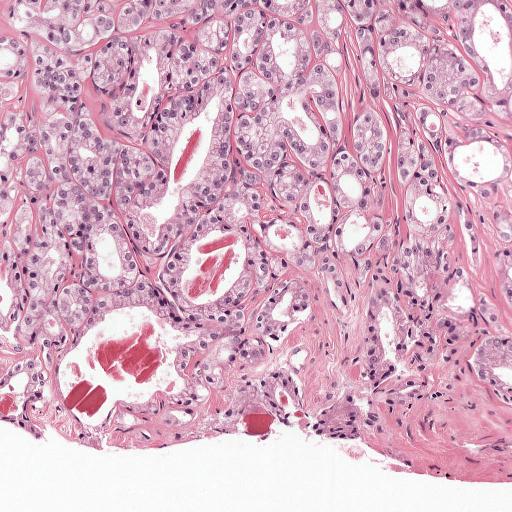}}
     }
     \subfloat[\textbf{\underline{Gleason 3}}\label{fig:fo}]{
         \centering
         \fbox{\includegraphics[width=0.13\linewidth]{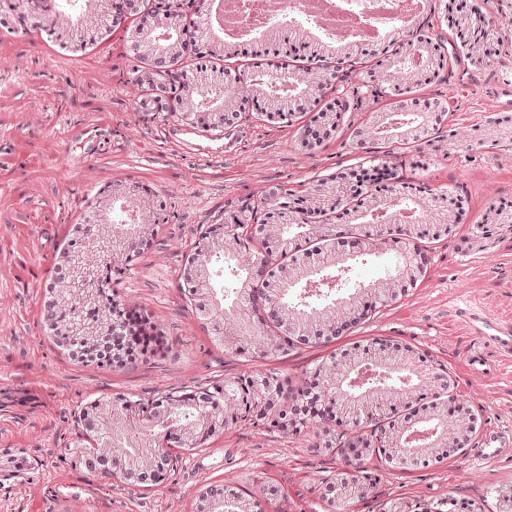}}
     }\\
     } %
     \caption{Acquired image patches. Each row shows the five patches with the highest acquisition score of the methods EN, BALD, and FocAL at the third acquisition step (after training with 800 labeled patches). Below each patch, we report the class and underline cancerous classes to highlight them. While the EN strategy favors ambiguous patches (like patches \ref{fig:enb} and \ref{fig:ene}) and BALD gets distracted by artifacts that do not contain cancerous tissue, the proposed FocAL method acquires informative patches that represent the cancerous classes well.}
        \label{fig:panda_acquisition}
\end{figure*}
In the diagnostic process, these uncertainties are important to identify unreliable predictions. A high epistemic uncertainty means that the model has to be further trained on this specific tissue type. A high aleatoric uncertainty indicates data ambiguities that might lead to a wrong class prediction. The OoD score shows that the indicated region contains artifacts or content that is substantially different from the learned data distribution. Low uncertainties - of all measures - indicate that the model probably made a correct classification in those areas.

\noindent 
\textbf{Acquisition} In Figure \ref{fig:panda_acquisition} we depict five patches with the highest acquisition score of the methods EN, BALD, and FocAL. They are taken from the third acquisition step but represent the general tendency that can be observed throughout the active learning process.
The proposed FocAL method avoids artifacts and ambiguities and acquires representative patches containing cancerous tissue. All five patches with the highest acquisition score contain either Gleason 3 or Gleason 4 in this acquisition step. Overall, in the complete AL process (big-Panda setup), the FocAL method acquired the highest number of Gleason 5 patches. It is the most severe grade and at the same time the most underrepresented one. In total 525 patches of Gleason 5 are acquired on average in the three runs (while EN acquired 403 and all other methods below 400 each). This shows, that the class imbalance was successfully addressed.
The EN acquisition assigns a high score for ambiguous patches that contain multiple different classes, like patches \ref{fig:enb} and \ref{fig:ene}. Both patches include glands of both Gleason 3 and Gleason 4, but for the classification task, only one label per patch is assigned, which is Gleason 4 for patch \ref{fig:enb} and Gleason 3 for patch \ref{fig:ene}. We argue that these ambiguities can slow down the labeling process because the pathologists take longer for the decision in comparison to the annotation of representative, non-ambiguous patches. 
The BALD method, which is commonly used and has shown impressive results on clean datasets \cite{gal_deep_2017}, fails to find informative patches. All five patches with the highest acquisition score contain artifacts and none of them contains cancerous tissue. Similar observations can be made for the MS and EP methods. All three methods (BALD, MS and EP) are highly 'distracted' by artifacts resulting in an acquired dataset in which many patches contain little or no tissue at all.

\begin{figure}[!t]
\centering
\includegraphics[width=0.6\linewidth]{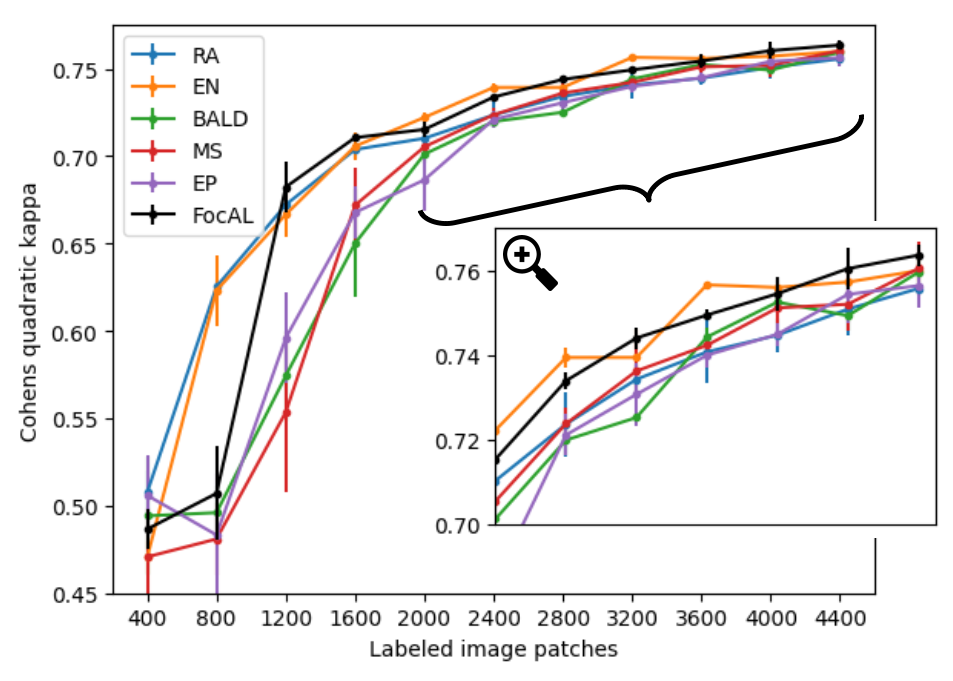}

\caption{Model Comparison on the Panda dataset (setting big-Panda). Several existing methods (BALD, MS, and EP) are below or on par with random acquisition (RA). This is the consequence of the acquisition of artifacts in the active learning process. The EN method acquires fewer artifacts, but more ambiguous images. This seems to be less harmful to the final result but leads to a more difficult labeling process and a less representative dataset. The proposed FocAL method actively avoids artifacts and ambiguities and reaches a satisfying performance with a final Cohen's quadratic kappa of $0.764$. }
\label{fig:panda_results_figure}
\end{figure}

\noindent 
\textbf{Model Comparison}
The proposed FocAL algorithm shows an overall strong performance as reported in Figure \ref{fig:panda_results_figure} with a final Cohen's quadratic kappa of $0.764$ with 4400 image patches corresponding to only $~0.69\%$ of all patches of the dataset. After the second acquisition step (with $1200$ labeled patches and more), the result is constantly better than RA, MS, EP, and BALD, because the acquisition of artifacts and ambiguities is actively avoided. As the FocAL acquisition selects representative patches of the classes, including many images of the most severe grade (Gleason 5) which is highly underrepresented, the created dataset is of high quality. Therefore, our algorithm has successfully addressed the challenges of histopathological labeling.

The methods BALD, MS, and EP, which are acquiring the highest number of artifacts, show a weak performance in the first acquisition steps. Their performance remains significantly below the baseline of random acquisition (RA) until 2400 patches are acquired. Afterwards, their performance is comparable to RA, but not significantly better. This confirms the previous findings of the MNIST experiments: these commonly applied acquisition methods perform well on clean datasets but artifacts have a major impact on their performance.

The EN method is less affected by the acquisition of artifacts but acquires more ambiguous images as already seen in Figure \ref{fig:panda_acquisition} and the MNIST experiment \ref{fig:mnist_acquisition}. Interestingly, this seems to not have a major impact on the performance for the Panda dataset and the final Cohen's quadratic kappas values are comparable to FocAL over several acquisitions. However, we want to stress some disadvantages of the EN method. For the MNIST experiment \ref{fig:mnist_results}, this method showed the worst performance and acquired the highest number of patches with ambiguities and artifacts. Therefore, it might not be suitable for other applications. Furthermore, the extensive acquisition of ambiguous images in the Panda dataset can lead to problems in the labeling process: the annotation by pathologists might take longer and the chances of wrong annotations are higher. Also, the final dataset might be less valuable because it mainly consists of edge cases of this specific model instead of representative patches of each class. The final Cohen's quadratic kappa value of EN is $0.759$ (FocAL: $0.764$).
 The supervised model with access to all $633.235$ patch labels reaches a Cohen's quadratic kappa of $0.841$ for this task. 

\noindent 
\textbf{Model Limitations}
Although FocAL successfully addresses the analyzed problems, there are some limitations worth discussing. We mimic the AL process with an already labeled dataset but it needs to be validated in a human study in the future. In practice, new problems but also advantages of FocAL might appear that are not notable in the experiments with an already labeled dataset. 
Additionally, we see some possibilities to further improve the model.
The aleatoric uncertainty is a widely adopted uncertainty measure for data uncertainty and overall captures ambiguities in the data well, but it sometimes shows false-positive values assigned to images close to the class boundary, as seen in the MNIST experiment (Figure \ref{fig:mnist_acquisition}). Indeed, we showed in the hyperparameter tuning for Panda (Figure \ref{fig:ablation_studies}) that the weight of the aleatoric uncertainty $\lambda_{al}$ must be carefully chosen. To further improve the model, a more precise uncertainty estimation for ambiguities could be a promising direction. 
Another limitation is that the model is currently trained from scratch for each acquisition step, as adapted from Gal \textit{et al.}\cite{gal_deep_2017}. Incrementally updating the model parameters at each step can reduce the overall training time, especially in the later acquisition steps.

\section{Conclusions} \label{sec:conclusions}
Our analysis of existing AL approaches for datasets with ambiguities and artifacts shows that these methods do not perform as expected. The widely used BALD algorithm, for example, acquires large amounts of images with artifacts, leading to performance that is often on par with or even below that of a random acquisition strategy. Furthermore, the resulting dataset is not a good representative of the classes of interest. Our proposed FocAL method addresses this issue by using precise uncertainty measures combined with OoD detection to avoid these ambiguities and artifacts while accounting for the class imbalance. In our experiments, we showed that each model component works as expected and that the overall results improve considerably. The acquired images are representative of the classes of interest and form a high-quality dataset. In the future, it would be interesting to analyze AL methods for other types of medical images, such as CT scans, dermatology images, or retinal images, with regard to artifacts, ambiguities, and class imbalance. It is likely that state-of-the-art methods such as BALD encounter similar problems, and the proposed FocAL method could provide a possible solution. In addition to these future applications, human studies are needed to validate the method in a real labeling process.

\bibliography{ms.bib}
\bibliographystyle{IEEEtran}

\end{document}